\documentclass[10pt,journal,compsoc]{IEEEtran}
\ifCLASSOPTIONcompsoc
  \usepackage[nocompress]{cite}
\else
  \usepackage{cite}
\fi

\ifCLASSINFOpdf
\else
\fi

\usepackage{graphicx}
\usepackage{amsmath}
\usepackage{amssymb}
\usepackage{booktabs}
\usepackage[english]{babel} 
\usepackage[T1]{fontenc}
\usepackage{array}
\usepackage{multirow}
\usepackage{amstext}
\PassOptionsToPackage{normalem}{ulem}
\usepackage{ulem}
\usepackage{bbding}
\usepackage{rotating}
\usepackage{colortbl}
\usepackage[dvipsnames]{xcolor}
\usepackage{enumitem}
\usepackage{cite}
\usepackage[colorlinks,linkcolor=red]{hyperref}

\begin{document}
%
\title{
\includegraphics[width=1cm]{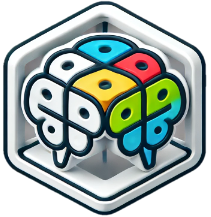}
MinD-3D++: Advancing fMRI-Based 3D Reconstruction with High-Quality Textured Mesh Generation and a Comprehensive Dataset
}

\author{
Jianxiong Gao, Yanwei Fu$^\dagger$, Yuqian Fu, Yun Wang, Xuelin Qian, Jianfeng Feng
\thanks{$^\dagger$: Corresponding author.}
\IEEEcompsocitemizethanks{
\IEEEcompsocthanksitem Jianxiong Gao, Yuqian Fu, Yun Wang, Xuelin Qian, Jianfeng Feng and Yanwei Fu are with Fudan University. Yanwei Fu is also with the Fudan ISTBI–ZJNU Algorithm Centre for Brain-Inspired Intelligence, Zhejiang Normal University, Jinhua, China, and the Shanghai Innovation Institute, Shanghai, China.
E-mail: jxgao22@m.fudan.edu.cn, yanweifu@fudan.edu.cn.
\IEEEcompsocthanksitem Dr. Yuqian Fu is now with ETH Zürich. And Dr. Xuelin Qian is now with Northwestern Polytechnical University.
\IEEEcompsocthanksitem This project is partly supported by STI 2030-Major Projects (No. 2022ZD0205300).
}
}

\markboth{Journal of \LaTeX\ Class Files,~Vol.~14, No.~8, August~2015}%
{Shell \MakeLowercase{\textit{et al.}}: Bare Demo of IEEEtran.cls for Computer Society Journals}
%


\IEEEtitleabstractindextext{%
\begin{abstract}

Reconstructing 3D visuals from functional Magnetic Resonance Imaging (fMRI) data, introduced as Recon3DMind, is of significant interest to both cognitive neuroscience and computer vision. To advance this task, we present the fMRI-3D dataset, which includes data from 15 participants and showcases a total of 4,768 3D objects. The dataset consists of two components: fMRI-Shape, previously introduced and available at \url{https://huggingface.co/datasets/Fudan-fMRI/fMRI-Shape}, and fMRI-Objaverse, proposed in this paper and available at \url{https://huggingface.co/datasets/Fudan-fMRI/fMRI-Objaverse}. fMRI-Objaverse includes data from 5 subjects, 4 of whom are also part of the core set in fMRI-Shape. Each subject views 3,142 3D objects across 117 categories, all accompanied by text captions. This significantly enhances the diversity and potential applications of the dataset.
Moreover, we propose \textbf{MinD-3D++}, a novel framework for decoding textured 3D visual information from fMRI signals. The framework evaluates the feasibility of not only reconstructing 3D objects from the human mind but also generating, for the first time, 3D textured meshes with detailed textures from fMRI data. We establish new benchmarks by designing metrics at the semantic, structural, and textured levels to evaluate model performance. Furthermore, we assess the model's effectiveness in out-of-distribution settings and analyze the attribution of the proposed 3D pari fMRI dataset in visual regions of interest (ROIs) in fMRI signals. Our experiments demonstrate that MinD-3D++ not only reconstructs 3D objects with high semantic and spatial accuracy but also provides deeper insights into how the human brain processes 3D visual information. Project page: \url{https://jianxgao.github.io/MinD-3D}.

\end{abstract}

}

\maketitle

\IEEEdisplaynontitleabstractindextext

%
\IEEEpeerreviewmaketitle

\begin{figure*}[t]
    \centering
    \includegraphics[width=\linewidth]{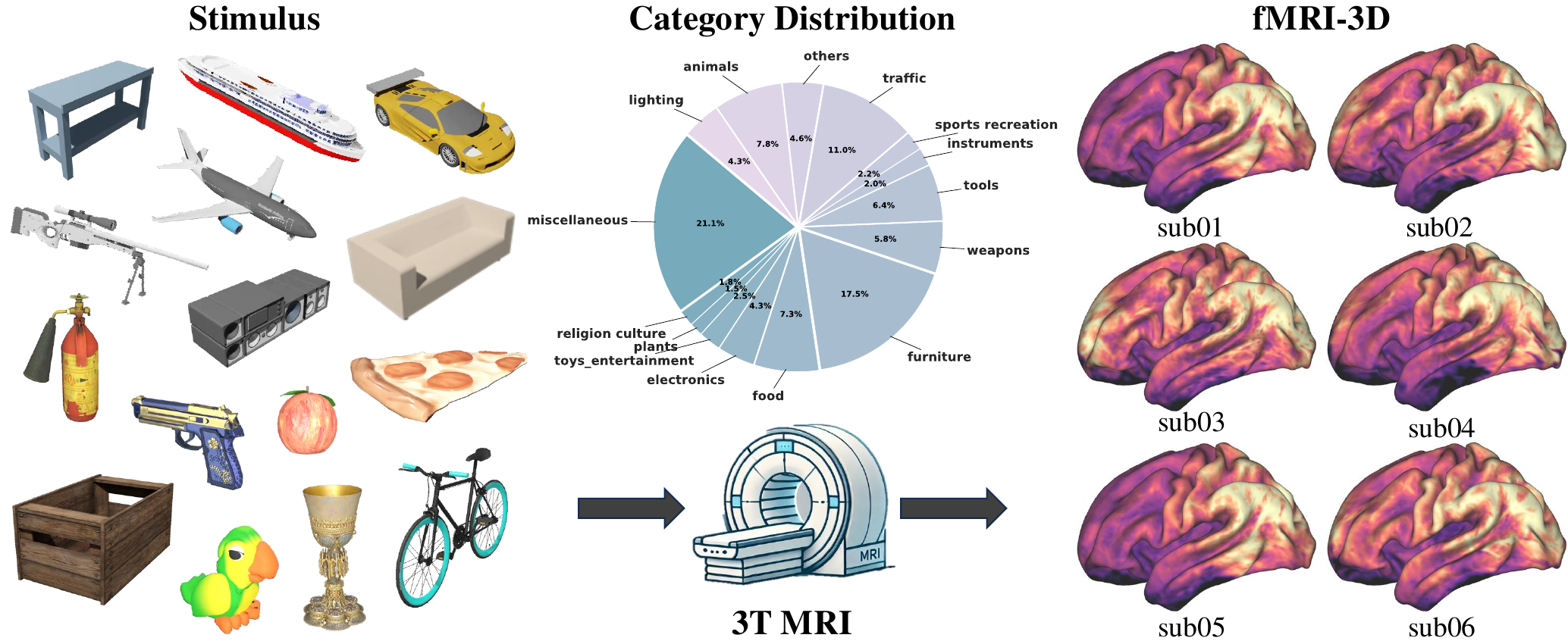}
\caption{\label{fig:teaser_all}
\textbf{Overview of our proposed fMRI-3D.} On the left, we show some 3D objects used as stimuli in our experiments; in the middle, a pie chart illustrates the distribution of 3D object categories, further highlighting the diversity of our data. On the right, fMRI signals from different subjects are displayed, showing varying neural responses to the same 3D object.}
\end{figure*}

\IEEEraisesectionheading{
\section{Introduction}
\label{sec:intro}
}

Functional Magnetic Resonance Imaging (fMRI), a kind of signal that can be obtained in a non-invasive way, could capture blood changes in the human brain induced by neuronal activity. Due to its relatively easy accessibility, fMRI has been commonly used to reflect visual activities.
Some recent studies~\cite{chen2023seeing,chen2023cinematic,scotti2023reconstructing,qian2023fmri,qian2023semantic} have successfully reconstructed high-quality images from fMRI signals by utilizing powerful generative models~\cite{rombach2021highresolution,chang2022maskgit}.
These approaches focus on extracting semantic features from fMRI signals, often requiring only semantic features to generate relevant high-quality images.

Existing methods primarily focus on reconstructing 2D visual information, but the human visual system extends far beyond merely processing flat images. 
The human brain appears capable of interpreting 2D projections in ways that support the perception of 3D structure~\cite{finlayson2017differential, wen2023identifying}.
This complex mechanism allows us to perceive the world in depth, recognizing attributes like size, distance, and spatial depth.
In contrast to previous studies, our research centers on modeling the brain's 3D visual capabilities. We introduce a new task, called \textbf{Recon3DMind} (\textbf{Recon}structing \textbf{3D} Objects from \textbf{Mind}), which leverages advanced computer vision techniques to decode and reconstruct the 3D visual information perceived by the brain from fMRI signals. This task goes beyond merely extracting semantic features, incorporating spatial and structural dimensions that are essential for a comprehensive understanding of 3D vision.

Several studies~\cite{groen2019scenes,10.1093/oso/9780190070557.003.0011,linton2023minimal} have demonstrated that the brain's mechanisms for 3D visual perception are significantly more intricate than those for 2D perception. This complexity is reflected in the distinct activation of brain regions during 3D visualization tasks~\cite{georgieva2009processing,jerath2015functional}. As a result, relying solely on semantic features is insufficient to fully model the brain's capacity for 3D spatial perception.
Effectively describing 3D objects requires taking into account not only their semantic features but also their shape and structural properties. For instance, two cars may appear identical when viewed head-on, yet differ greatly in length when viewed from the side. This example underscores the importance of capturing the full range of spatial and structural features to authentically represent 3D objects.
Accordingly, our work seeks to advance the modeling of human 3D perception by developing an enhanced fMRI feature extractor. This extractor is designed to capture semantic elements, spatial structures and other 3D-specific characteristics from fMRI signals. This approach aims to enable a more complete and accurate reconstruction of 3D visual information.

\begin{figure*}[t]
    \centering
    \includegraphics[width=\linewidth]{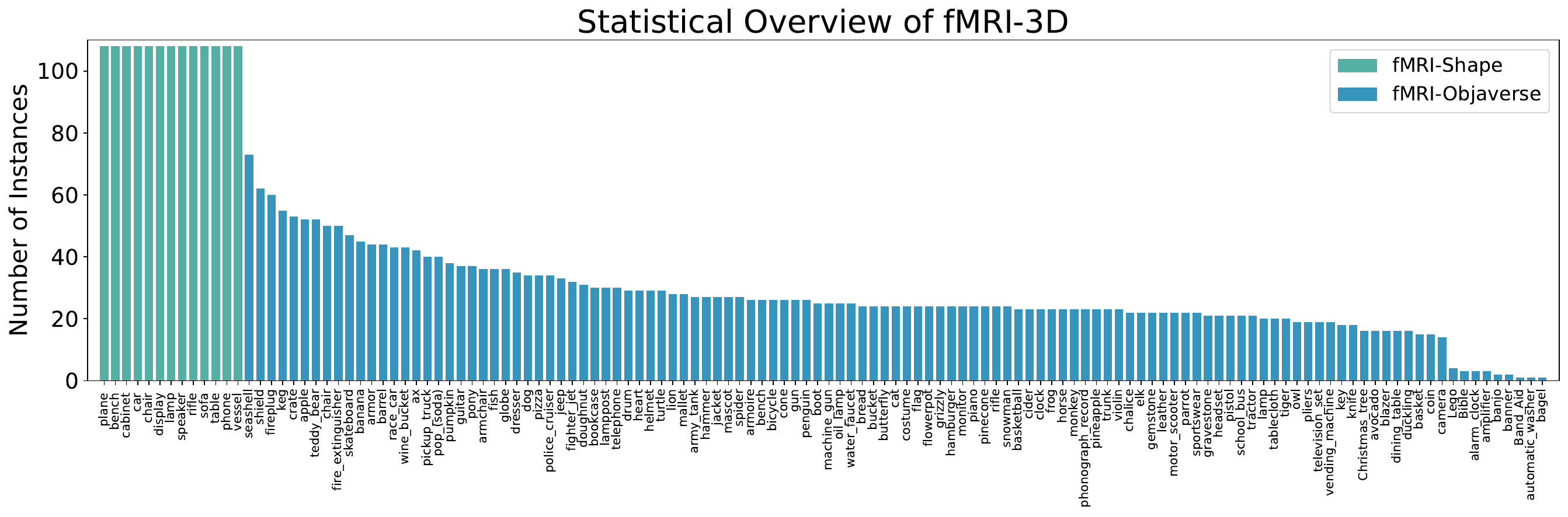}
\caption{
\textbf{Statistical Overview of Proposed fMRI-3D.} It displays the number of instances for each object category in the Core Set of the fMRI-Shape and fMRI-Objaverse datasets. fMRI-Objaverse significantly complements fMRI-Shape by offering a wider range of categories
\label{fig:fmriobj}
}
\end{figure*}

In our conference work~\cite{gao2023mind3d}, we introduced the fMRI-Shape dataset to tackle the significant challenge of the lack of datasets pairing fMRI data with 3D visuals for this complex task. The dataset comprises data from 14 participants and 1,624 3D objects. However, the Core set was limited to just 13 categories of 3D objects. To address this limitation in category diversity and to expand the number of objects, we propose fMRI-Objaverse, which includes data from 5 participants and 3,142 3D objects across 117 categories, accompanied by text captions. Notably, it shares 4 participants with the Core set of fMRI-Shape, significantly enhancing the diversity of fMRI-Shape, as shown in Fig.~\ref{fig:fmriobj}. We collectively refer to these two datasets as \textbf{fMRI-3D}, aiming to support various experimental setups and further promote research within the community.

\begin{figure*}[ht]
    \centering
    \includegraphics[width=0.98\linewidth]{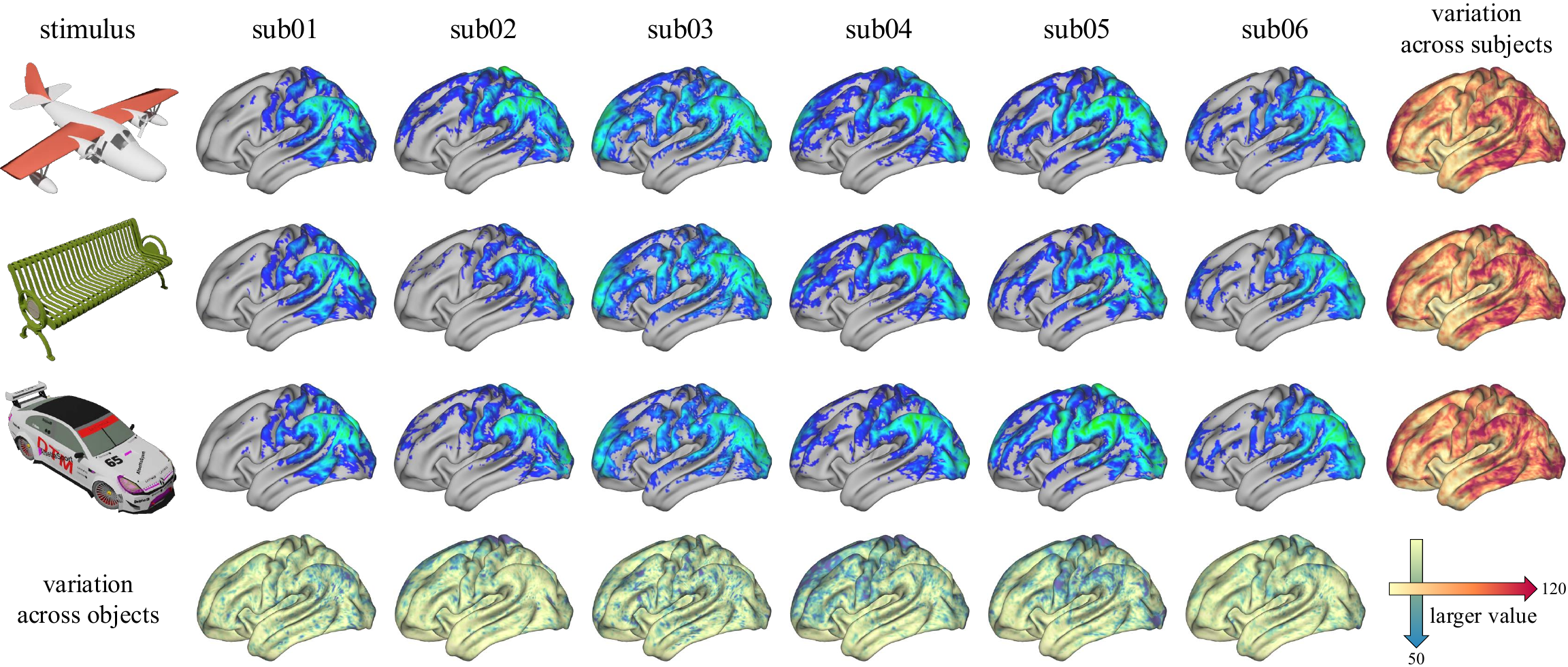}
\caption{
\textbf{Individual differences in brain activation patterns within the fMRI-3D dataset.} In our dataset fMRI-3D, the variation in brain activity across different participants viewing the same object is greater than the variation when the same participant views different objects. Red and blue regions represent areas with higher values for variation across subjects and variation across objects, respectively.
\label{fig:fmri}
}
\end{figure*}

During fMRI data collection, we present 3D objects through 360-degree view videos, providing comprehensive visualizations that stimulate the brain’s perception of 3D objects and facilitate the collection of high-quality data. In our approach, participants watch 360-degree videos of stationary 3D objects from ShapeNet~\cite{chang2015shapenet} and Objaverse~\cite{objaverse}, where a rotating camera completes a full orbit around each object, offering a complete view from all angles. This method ensures detailed and accurate capture of fMRI signals, as participants engage with the objects, allowing for the full range of spatial features to be recorded. As shown in Fig.~\ref{fig:fmri}, we also analyze the variation in fMRI data across both subjects and objects. Interestingly, the variation across subjects is even greater than that across objects. After careful preprocessing, these recordings are transformed into multi-frame fMRI signals, resulting in a rich dataset for detailed analysis. The complexities and specific features of the fMRI-3D dataset will be discussed further in Sec.~\ref{sec:3}.

\begin{figure*}[t]
    \centering
    \includegraphics[width=0.98\linewidth]{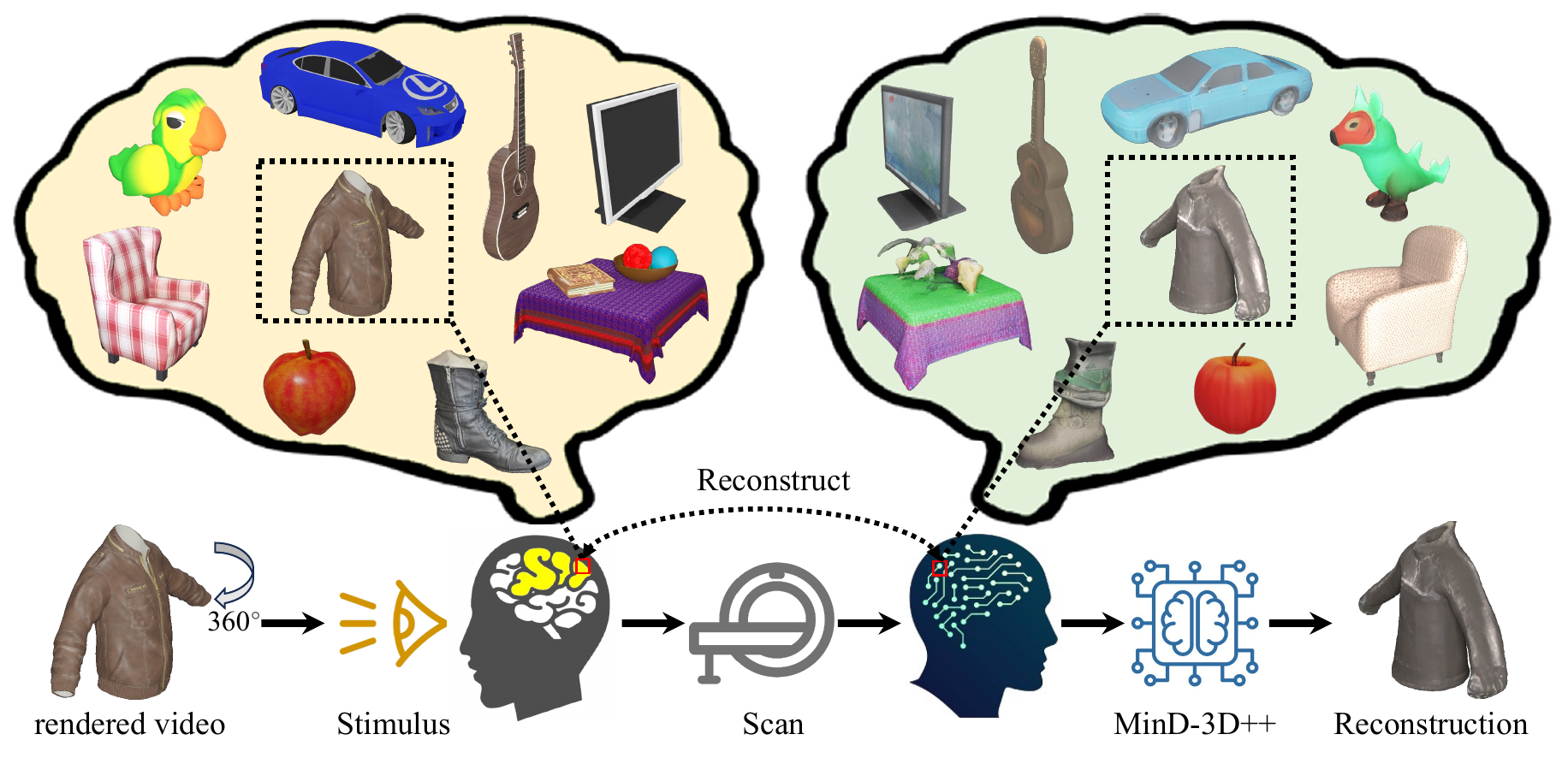}
\vskip -0.1in
\caption{\label{fig:teaser}
\textbf{Pipeline of Recon3DMind task}, showcasing the fMRI-3D dataset collection process, where participants observe 360-degree videos of 3D objects, and the MinD-3D++ framework for reconstructing textured 3D objects from fMRI signals.
}
\end{figure*}

Leveraging our carefully curated fMRI-3D dataset, we introduce \textbf{MinD-3D++}, a novel framework which generates textured 3D visual stimuli directly from fMRI signals for the first time. The framework comprises two main steps: (1) extracting features from multi-frame fMRI signals, and (2) generating corresponding multi-view images and subsequently synthesizing 3D objects.

During feature extraction, we first use a transformer-based encoder~\cite{qian2023semantic} pre-trained on the NSD dataset~\cite{allen2022massive} to extract spatial features from the fMRI data. These features are then aggregated across multiple frames through a feature aggregation module. To maintain biological relevance and ensure the effectiveness of the extracted features, we align them with both the visual and textual representations of the corresponding objects. This alignment is achieved by applying a contrastive learning loss on the class token within the encoder’s blocks.

Once the features are aligned, we leverage the strong generative capabilities of diffusion models to accurately capture object appearance. Specifically, we adopt a multi-view diffusion model in which the extracted features and class token serve as conditional inputs to a pre-trained diffusion model. Through LoRA-based fine-tuning of the attention layers, we enable the model to incorporate fMRI-derived features. This process yields six-view images that faithfully reflect the original 3D stimuli. Finally, using a pre-trained model, we synthesize a textured 3D mesh from these multi-view images, completing the end-to-end generation of 3D visual stimuli.

To evaluate the effectiveness of our model, we design new benchmarks that measure performance across semantic, structural, and textured levels. These metrics provide a comprehensive assessment of the model’s ability to generate 3D representations that are both structurally and semantically accurate. We test our model in both standard and out-of-distribution settings, where it consistently outperforms baseline models. Additionally, we conduct in-depth analyses of the fMRI-3D dataset and the features extracted by MinD-3D. These analyses explore how the brain perceives different angles, objects, and semantic information within specific regions of interest (ROIs). We further validate the biological relevance of our model's features by correlating them with brain regions, demonstrating that the generated representations align with the brain's visual information processing.

This paper builds upon our preliminary conference work, and we summarize the key contributions as follows:

\begin{itemize}[leftmargin=*,itemsep=0pt,topsep=0pt,parsep=0pt] 
\item We introduce \textbf{fMRI-Objaverse}, a large-scale extension of the original dataset. Together with fMRI-Shape, we collectively refer to these datasets as fMRI-3D, which is designed to support various experimental setups and advance research in the field. 
\item We propose an improved framework, \textbf{MinD-3D++}, for the first time, capable of reconstructing textured meshes from fMRI signals, representing a significant advancement in decoding 3D representations from human mind.
\item We establish a comprehensive benchmark for the task of 3D visual reconstruction from human brain data. 
\item We conduct extensive experiments to analyze the contributions of our proposed dataset to decoding fMRI signals, further validating the effectiveness of \textbf{MinD-3D++}. 
\end{itemize}

\section{Related work}
\label{sec:Related_work}

\begin{figure}[tb]
    \centering
    \includegraphics[width=0.8\linewidth]{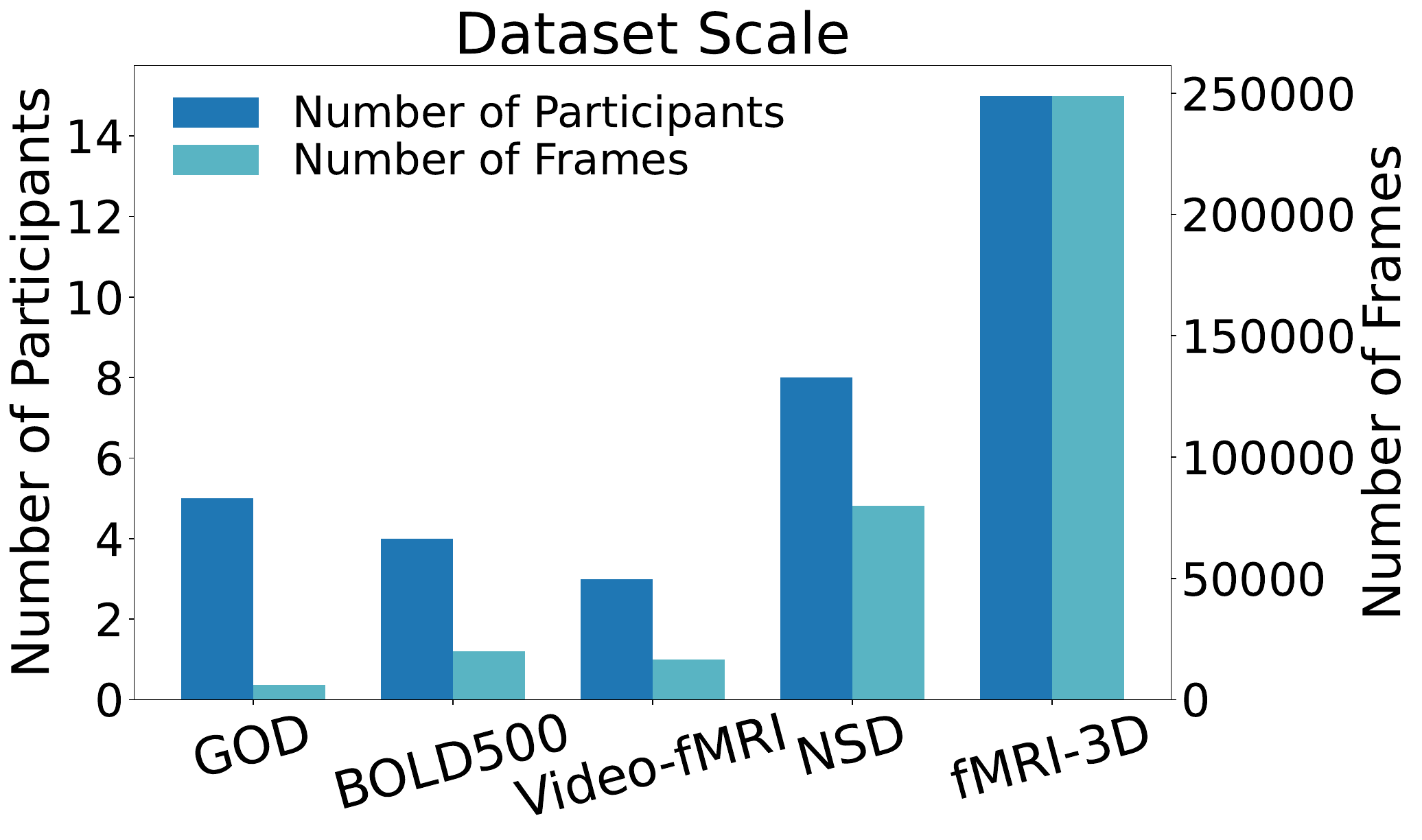}
    \caption{\textbf{Comparing fMRI-3D with other 2D fMRI datasets.} As the first 3D fMRI dataset, fMRI-3D features a larger number of participants and frames, providing ample support for experiments in our proposed novel task and further research.
    \label{fig:dataset_scale}
    }
\end{figure}

\subsection{fMRI Decoding Methods}

Current fMRI decoding methods primarily focus on reconstructing the vision perception in a 2D format, such as the images or videos perceived by humans. This is a challenging task, as it involves extracting relevant features from fMRI signals with precision to recreate accurate 2D representations. Deep learning methods, known for their impressive capabilities, are particularly suited to address this challenge. Initial successes in this area have been demonstrated by earlier methods~\cite{horikawa2017generic,wen2018neural,shen2019deep}. Subsequent studies~\cite{shen2019end, du2019brain} have shown that generative models are particularly effective for these tasks, leading to the employment of various diffusion models~\cite{chen2023seeing, chen2023cinematic, scotti2023reconstructing, sun2024contrast} as decoders to reconstruct visual scenes, achieving remarkable results. However, these studies have been limited to 2D visual representations and the related vision ROIs. In this paper, we aim to extend the scope of fMRI visual decoding to 3D representations, involving more vision ROIs. Our goal is to directly reconstruct 3D objects from fMRI signals. To accomplish this, we propose a new framework that employs a transformer-based feature encoder for extraction and aggregation. This framework translates neural space data into visual space and utilizes a powerful 3D decoder to reconstruct the 3D object, leveraging features from the visual space.

\subsection{Diffusion Models}
Diffusion models~\cite{ho2020denoising,song2020denoising} are exceptional generative tools for both pixel and feature generation. As a variant, the latent diffusion model~\cite{rombach2021highresolution}, equipped with an autoencoder, compresses images into lower-dimensional latent features, thereby generating a compressed version of the data rather than directly generating the data itself. Dit~\cite{peebles2023scalable} replaces the backbone of diffusion models with transformers, which will improve the performance and scalability of these models. This approach, operating in the latent space, significantly reduces computational requirements and enables the generation of higher-quality images with enhanced details in the latent space. In this paper, we aim to leverage the potent feature-generation capabilities of diffusion models to generate visual features based on fMRI features. To achieve this, we adapt a transformer-based diffusion model, focusing solely on its latent component. The conditional information driving the model is derived from the fMRI features.

\subsection{3D Generation}
3D generation can be accomplished through various methods~\cite{qian2024pushing,tang2024lgm,liu2024mirrorgaussian}. Some methods~\cite{tang2023dreamgaussian, tang2024lgm} employ 3D Gaussian splatting~\cite{kerbl3Dgaussians} for this purpose.      
Other studies~\cite{wang2023imagedream, ye2024dreamreward} utilize diffusion models to generate multi-view representations of objects, subsequently constructing 3D models. Additionally, traditional and direct approaches leverage autoregressive methods~\cite{cheng2022autoregressive, ibing2023octree, qian2024pushing} for 3D object generation.
In our study, we adapt Argus~\cite{qian2024pushing}, a robust 3D generative model with several transformer layers, as our decoder to generate 3D objects from fMRI data. This approach integrates visual features generated by the preceding diffusion module. These visual features serve as conditional embeddings for Argus. This synergistic integration aims to enhance the model's ability to accurately reconstruct 3D objects from complex brain activity.

\section{Curated Dataset}
\label{sec:3}
In this section, we detail the procedures for collecting the proposed fMRI-3D dataset, which consists of two components: fMRI-Shape and fMRI-Objaverse. The scale of fMRI-3D is compared to other benchmark datasets, including NSD~\cite{allen2022massive}, BOLD5000~\cite{chang2019bold5000}, GOD~\cite{horikawa2017generic}, and Video-fMRI~\cite{wen2018neural}, as illustrated in Fig.~\ref{fig:dataset_scale}. Specific details about fMRI-Shape and fMRI-Objaverse are provided in Tab.~\ref{tab:detail_about_dataset}. For all experiments, written informed consent was obtained from each participant, and the study was approved by the ethical review board.
To better illustrate brain activation patterns and demonstrate the utility of the fMRI-3D dataset, we analyze and visualize responses to three distinct objects across six subjects, as shown in Fig.~\ref{fig:fmri}. Note that only voxels with activation levels above the 50th percentile are displayed. We also compute the variation across subjects and objects, with red and blue regions indicating higher activation values, respectively, reflecting areas in the human brain sensitive to the stimuli. This visualization highlights significant individual differences in brain activation across subjects, which are more pronounced than the variations in responses to different objects. These findings emphasize the inherent challenges and underscore the importance of the 
Across-Person (AP) and Across-Person \& Across-Class (APAC) settings.
All participants~\footnote{\textbf{Acknowledgement:} We thank all volunteers for participating in our fMRI experiments. We are especially grateful to Yansong Bai, Ruiying Chen, Yachen Gao, Junwei Liu, Yunheng Li, Tiran Li, Tongying Pan, Yao Xiao, and Yuxi Zheng for their additional support during the data collection process.} had normal or corrected-to-normal vision. The fMRI-3D dataset will be made publicly available to support further research in Recon3DMind.

\subsection{fMRI-Shape}

\begin{figure*}[t]
    \centering
    \includegraphics[width=0.9\linewidth]{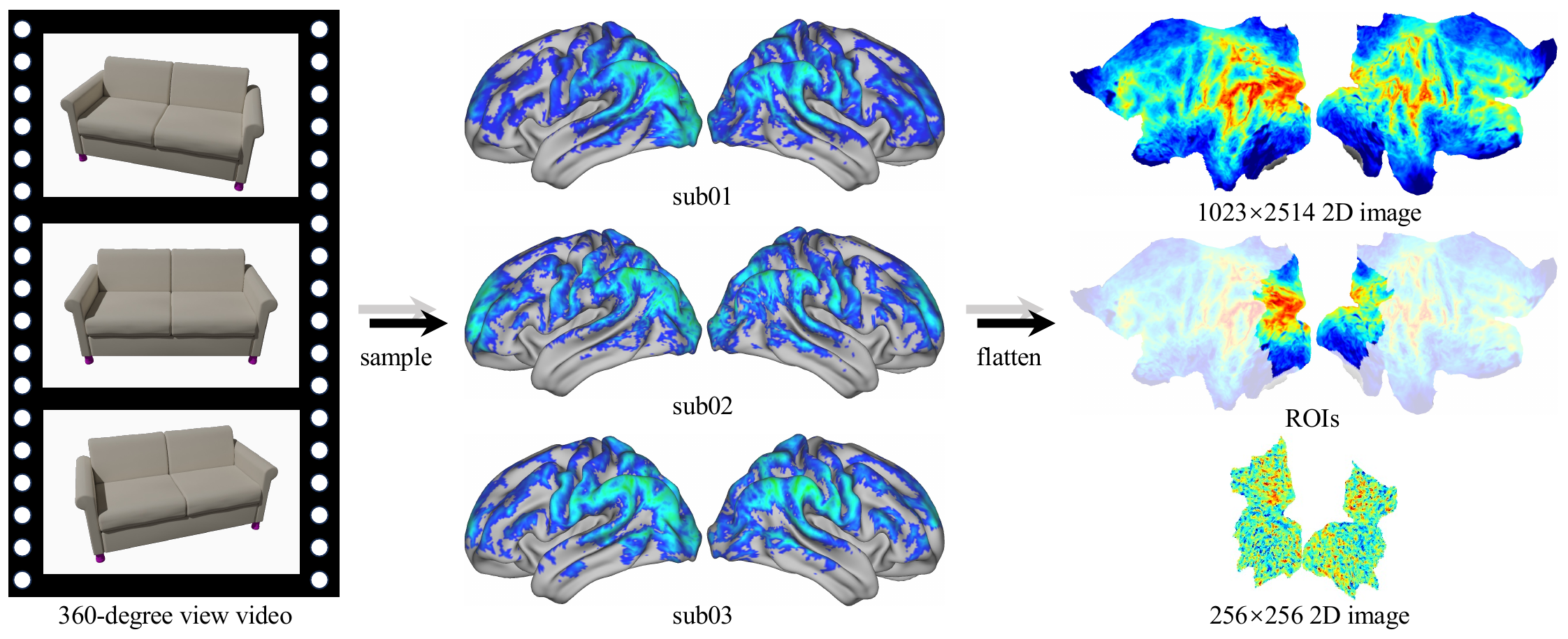}
\caption{
\textbf{Overview of the fMRI-3D Acquisition Process.} Initially, we render each object into an 8-second long video, showcasing a 360-degree view. Subsequent fMRI signal capture is performed in video format, followed by data processing with fMRIPrep to convert signals from 32k\_fs\_LR surface space into 2D images of dimensions 1023 $\times$ 2514. \textbf{Individual differences} observed in the dataset, as highlighted in the middle part, underscore the challenges in generalizing these findings. On the rights, regions of interest (ROIs) are transformed into 256 $\times$ 256 image.
}
\label{fig:dataset_example}
\end{figure*}

FMRI-Shape contains data from 14 participants who were unaware of the objectives of the work.
To ensure diversity in the dataset, the 3D objects were sourced from ShapeNetCore~\cite{chang2015shapenet}, which includes 55 object categories. We employed the rendering technique from Zero123~\cite{liu2023zero1to3} to render 192 images using Blender and generated 8-second videos at 24 fps for each object.
These videos depict the 3D objects rotating 360 degrees at a 60-degree pitch angle, as illustrated in Fig.~\ref{fig:dataset_example}. The dataset is available for download at: 
\url{https://huggingface.co/datasets/Fudan-fMRI/fMRI-Shape}.

\noindent \textbf{1) Core Set:}  
The core set of fMRI-Shape includes data from 8 participants (4 males and 4 females, aged 21 to 29, Participants No. 1-8). A total of 1,404 objects were selected from 13 commonly used categories in 3D reconstruction literature~\cite{chen2019learning, sun2020pointgrow, ibing20213d} within ShapeNetCore. For each category, 100 objects were used for training and 8 for testing, resulting in 108 objects per category. For more details, please refer to our conference version~\cite{gao2023mind3d}.

\noindent \textbf{2) Across-Person Set (AP Set):}  
The AP Set was designed for Out-of-Distribution (OOD) testing and includes fMRI data from 2 participants (1 male aged 24 and 1 female aged 26, Participants No. 9 and 10) who viewed the test objects from the Core set.

\noindent \textbf{3) Across-Person \& Across-Class Set (APAC Set):}  
The APAC Set presents a more challenging OOD test compared to the AP Set. It includes data from 4 participants (2 males and 2 females, aged 22 to 26, Participants No. 11-14). For this set, we randomly selected 4 objects from each of the 55 categories in ShapeNetCore, distinct from those in the Core set, resulting in a total of 220 objects. 

As illustrated in Fig.~\ref{fig:fmri}, and the middle part of Fig.~\ref{fig:dataset_example}, individual differences among participants pose significant challenges for generalization. The AP and APAC sets are crucial for OOD testing and will serve as important benchmarks for assessing the generalization capability of 3D decoding models.

\begin{table}[tb]
    \centering
    \caption{
        \textbf{Details of fMRI-3D Dataset.} The large-scale dataset ensures a balanced representation of male and female participants and includes both the fMRI-Shape and fMRI-Objaverse datasets. The fMRI-Shape dataset comprises three distinct subsets: the Core set, the Across-Person set (AP Set), and the Across-Person \& Across-Class set (APAC Set), which support standard and out-of-distribution (OOD) experimental settings. The latter two subsets are designed to facilitate model generalization evaluations. The fMRI-Objaverse dataset extends fMRI-Shape by featuring four of the same participants viewing a wider variety of 3D objects from Objaverse, accompanied by text captions.
        \label{tab:detail_about_dataset}
    }
    \begin{tabular}{lccccc}
    \toprule
      &  P. & Male/Female & Category  & Obj & Frames \\
    \midrule
    \midrule
    fMRI-Shape & 14 & 7/7 & 55 & 1624 & 123200 \\
    \midrule
    \quad Core Set   & 8    &  4/4   & 13   &  1404  & 14040\\
    \quad AP Set & 2    &  1/1   & 13   &  104   & 1040 \\
    \quad APAC Set & 4   &  2/2   & 55   &  220 &  2200 \\
    \midrule
    fMRI-Objaverse & 5   &  2/3   & 117   &  3142 &  125680\\
    \midrule
    fMRI-3D (Total) & 15 & 7/8 & 172 & 4768 &  248880 \\
    \bottomrule
    \end{tabular}
\end{table}

\subsection{fMRI-Objaverse}

\begin{figure*}[t]
    \centering
    \includegraphics[width=0.95\linewidth]{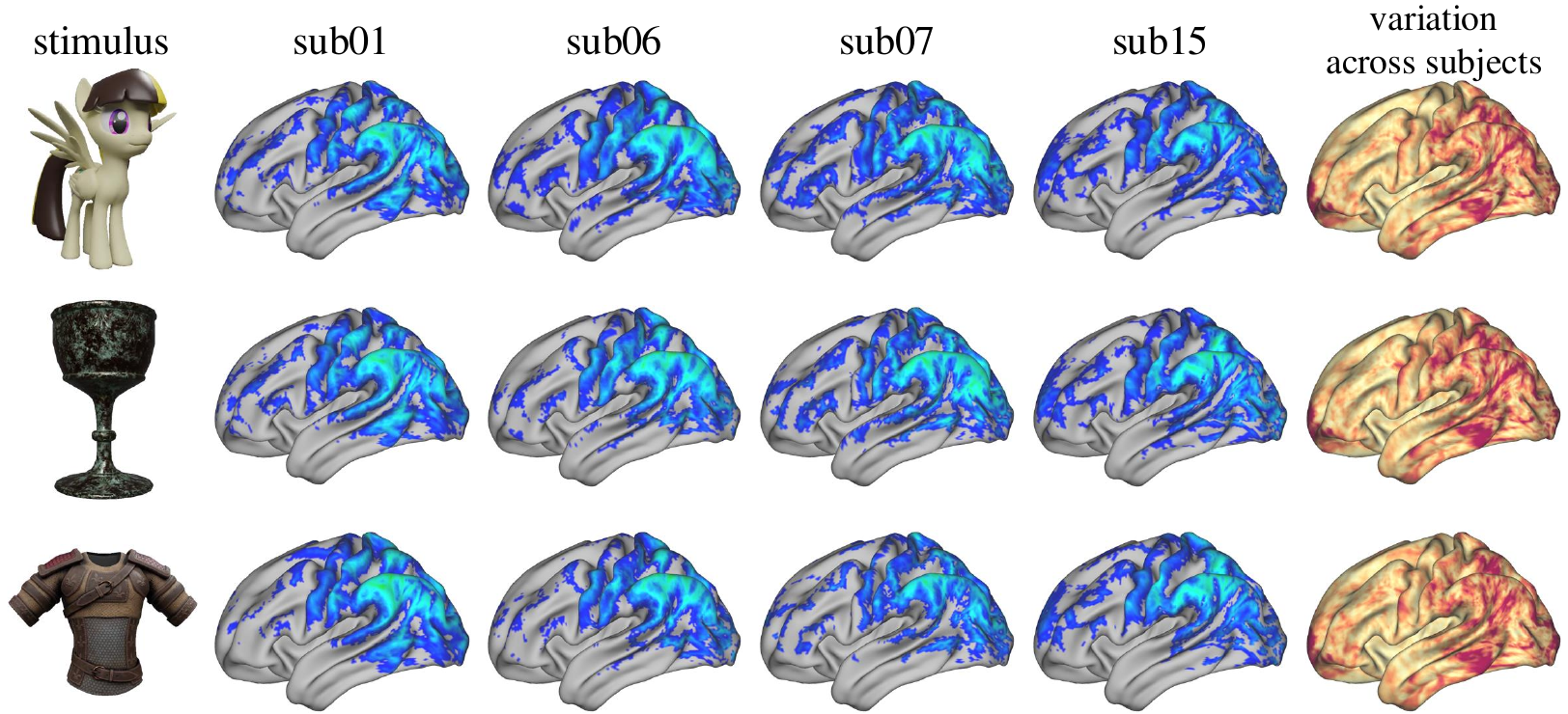}
\caption{
\textbf{Individual differences in brain activation patterns within the fMRI-Objaverse dataset.} In our extensive fMRI-Objaverse dataset, the variation in brain activity across different participants viewing the same object is highly pronounced. Red regions represent areas with higher levels of variation across subjects. This again indicates the essential challenges of our proposed task.
\label{fig:fmri_obj}}
\end{figure*}

FMRI-Objaverse, as partially shown in Fig.~\ref{fig:fmri_obj}, includes data from five participants, all of whom were unfamiliar with the study. Four participants (2 males and 2 females, aged 22 to 26, identified as Nos. 1, 6, 7, and 8) overlap with the core fMRI-Shape dataset, providing an important extension in terms of diversity and scale. Additionally, the fifth participant (No. 15), a 22-year-old female, was included to further expand the dataset.
To enhance our dataset, we selected 3,142 objects from the top 117 object categories in Objaverse~\cite{objaverse}, based on a subset filtered by LGM~\cite{tang2024lgm} and enriched with text descriptions from Cap3D~\cite{luo2023scalable}. Unlike in fMRI-Shape, each 3D object in this dataset was rendered into 384 frames, generating a 6.4-second video at 48 fps using Blender.
Each participant spent approximately 8 hours in experimental sessions, divided into 53 sessions. During each session, participants viewed 60 videos in a randomized order, except for the last session, with 1.6-second rest intervals between each pair of objects. To prevent low-quality data due to visual fatigue, we randomly reversed the rotation direction for 40\% of the selected objects. All objects were presented once to each participant. (Note: Our MRI machine samples data every 800ms, so we selected 6.4-second videos with a 1.6-second rest period between them.)
This extension supports further multimodal experiments and applications. The objects in Objaverse contain more details and a wider variety of categories, and the higher fps videos present stronger visual effects, posing a significant challenge for reconstructing them in fMRI-Objaverse. The dataset is available for download at: \url{https://huggingface.co/datasets/Fudan-fMRI/fMRI-Objaverse}.

\subsection{Data Acquisition and Preprocessing}
The T1 and fMRI data were acquired in a 3T scanner and a 32-channel RF head coil. T1-weighted data were scanned using MPRAGE sequence (0.8-mm isotropic resolution, TR=2500ms, TE=2.22ms, flip angle $8^{\circ}$). Functional data were scanned using gradient-echo EPI at 2-mm isotropic resolution with whole-brain coverage (TR=800ms, TE=37ms, flip angle $52^{\circ}$, multi-band acceleration factor 8). The sampling frequency of the 3T scanner is 1.25Hz, so each video segment corresponds to a total of 10 frames of task-state fMRI signals.

Stimuli were presented using an LCD screen ($8^{\circ}\times 8^{\circ}$) positioned at the head of the scanner bed. Participants viewed the monitor via a mirror mounted on the RF coil and fixated a red central dot (0.4° × 0.4°). 

Preprocessing was performed using fMRIPrep~\cite{fmriprep1,fmriprep2}. Following~\cite{qian2023fmri}, the preprocessed functional data in 32k\_fs\_LR surface space were converted into 2D images and utilized for further analysis. Given the delay of the BOLD signal by 6 seconds, we applied z-scoring to the data points across every vertex within each run, incorporating a 6.4-second lag. These normalized values were then projected onto 1023 × 2514 pixel 2D images using pycortex. For analysis, Regions of Interest (ROIs) were selected from the Human Connectome Project Multi-Modal Parcellation (HCP-MMP) atlas in the 32k\_fs\_LR space. These ROIs included areas such as ``V1, V2, V3, V3A, V3B, V3CD, V4, LO1, LO2, LO3, PIT, V4t, V6, V6A, V7, V8, PH, FFC, IP0, MT, MST, FST, VVC, VMV1, VMV2, VMV3, PHA1, PHA2, PHA3, TE2p''. Subsequently, the ROIs were converted into a 256 × 256 image, as illustrated in the right part of Fig.~\ref{fig:dataset_example}.

\begin{figure*}[t]
    \centering
    \includegraphics[width=0.98\linewidth]{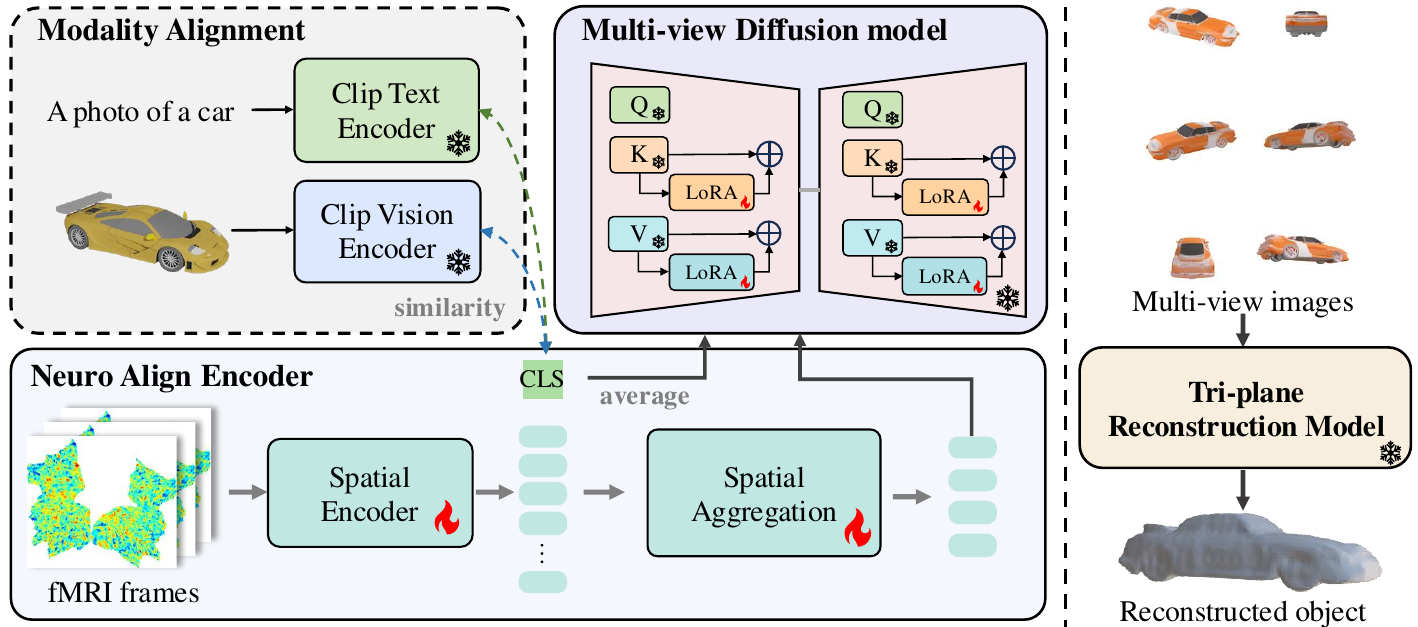}
\caption{
\textbf{Overview of the MinD-3D++ Framework.} Our improved approach fully leverages the class token in the Neuro-align Encoder, incorporating image and text contrastive learning to extract features from fMRI frames. We then use the LoRA fine-tuning method to refine the QV linear layers in the attention of the pre-trained multi-view diffusion model, improving the generation of detailed and textured representations of the target object in a multi-view presentation. Finally, we use the tri-plane reconstruction model to obtain the textured mesh.
}
\label{fig:framework_new}
\end{figure*}

\section{Problem Setup}
Recon3DMind tackles a critical challenge in cognitive neuroscience: developing computational models that can accurately interpret and reconstruct the brain's 3D visual comprehension. This endeavor not only bridges the gap between cognitive neuroscience and computer vision but also has the potential to advance the latter field in unprecedented ways. In this paper, we focus on the specifics of fMRI-based 3D reconstruction, providing detailed definitions and formulas that underpin our approach.

We begin with the acquisition of a multi-frame fMRI signal, denoted as $\{F\}$, where $|F|=n$ (with $n=8$ or $n=10$). These signals correspond to both a 3D object mesh $\Psi$, and a video $\{V\}$, with $|V|=k$ (where $k=192$ or $k=384$ frames), which the subject observes. The task requires an efficient encoder, $E$, capable of extracting both spatial structural and semantic features from the fMRI signal. It is important to note that while a single frame of fMRI data is sufficient to extract semantic information for 2D image reconstruction, reconstructing 3D structures requires additional spatial structural features. Therefore, multiple frames of fMRI data are input to capture these comprehensive spatial features from the spatio-temporal signals. 

After feature extraction, a powerful decoder is used to reconstruct the original 3D mesh $\Psi$, based on the extracted feature $f$:  $\Psi = D(f)$. Thus, our model can be succinctly described as $ M = \{E, D\}$, where the transformation is represented as $\Psi = M(F)$.

To effectively implement this model, it is crucial to leverage both the fMRI signals $\{F\}$ and the corresponding video $\{V\}$ to train the model $M$. For this, we propose a three-stage, innovative, and efficient framework. Each stage is carefully designed to capture different aspects of the fMRI data and the associated visual stimuli, ensuring a comprehensive and accurate 3D reconstruction from the complex neural signals. This process not only pushes the boundaries of current computer vision techniques but also provides valuable insights into how the human brain processes 3D spatial information.


\section{Method}

\subsection{Preliminary}
\label{limitofmind3d}

\begin{figure}[t]
    \centering
    \includegraphics[width=\linewidth]{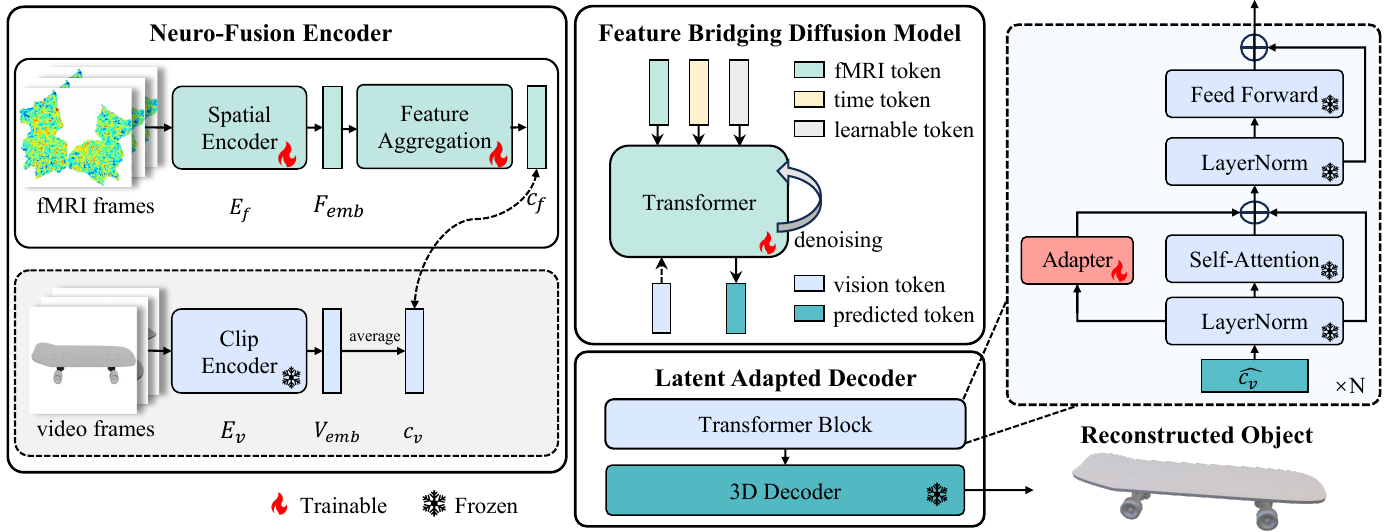}
    \caption{
    \textbf{Pipeline of MinD-3D.} The conference version of our model uses a three-stage framework to reconstruct 3D meshes from fMRI.
    }
    \label{fig:framework_old}
\end{figure}

Our conference model, MinD-3D~\cite{gao2023mind3d}, shown in Fig.~\ref{fig:framework_old}, demonstrates the feasibility of reconstructing 3D meshes from human brain data. MinD-3D combines a neuro-fusion encoder for extracting features from fMRI frames, a feature-bridged diffusion model for generating visual features from these fMRI signals, and a latent-adapted decoder based on the Argus 3D shape generator for reconstructing 3D objects. This integrated system effectively aligns and translates brain signals into accurate 3D visual representations.

Despite the success of MinD-3D, one important aspect remains unexplored: the reconstruction of 3D objects with texture. Inspired by the strong appearance-generation capabilities of 2D diffusion models, we propose a novel approach for exploring textural reconstruction. This approach involves enhancing the encoder and adopting a more robust, multi-view-based 3D generation model. Our goal is to use the diffusion model to generate both the appearance and texture of 3D objects. This represents the first attempt to generate textured meshes from fMRI data and expands the range of object categories that can be decoded from fMRI into 3D representations. Therefore, we introduce the improved framework, \textbf{MinD-3D++}, as illustrated in Fig.~\ref{fig:framework_new}.

\begin{figure*}[t]
\centering
\includegraphics[width=0.95\linewidth]{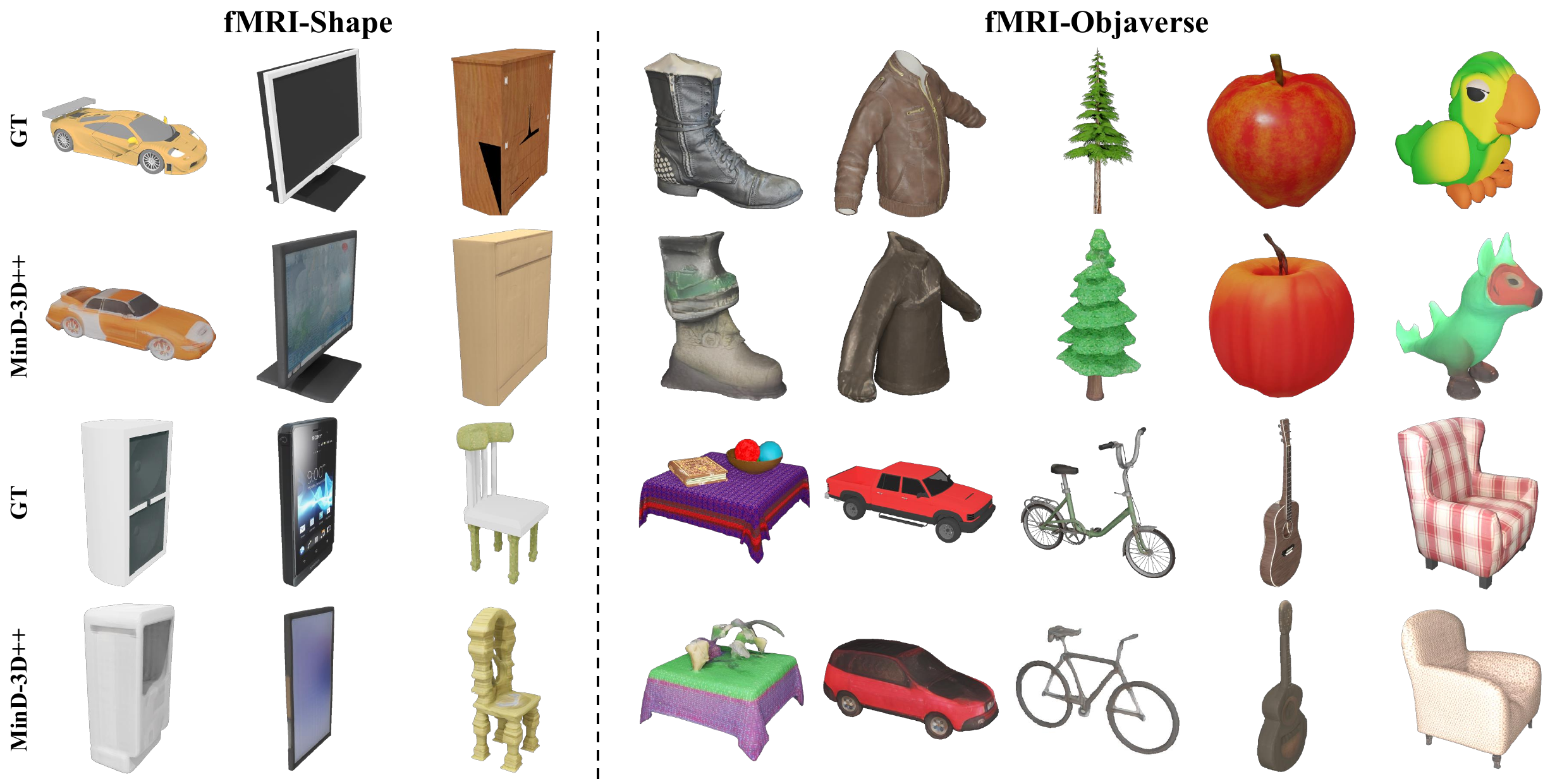}
\caption{
\textbf{Qualitative Results of MinD-3D++ on fMRI-3D.} To demonstrate the effectiveness of our MinD-3D++ model, we present the reconstruction of textured meshes, with ground truth (GT), from both fMRI-Shape and fMRI-Objaverse.
\label{fig:mind3dplus_vis}
}
\end{figure*}

\subsection{Neuro Align Encoder}
We used the same encoder architecture as in MinD-3D. In the original work LEA~\cite{qian2023semantic}, the encoder utilized a class token that was trained to reconstruct the complete raw fMRI signal by masking all spatial tokens. This class token is informative and powerful, and in this improved model, we focus on utilizing the class token and apply contrastive learning to it, using it as part of the conditional information for the subsequent diffusion model.

We still process each fMRI frame in parallel to obtain both the spatial fMRI embeddings and the class tokens:
\begin{equation}
    T_c^i, F_{emb}^i = E_{f}(F^i),
\end{equation}
where $i \in {1, 2, \dots, N}$.

For spatial information from the fMRI signals, we still employ an aggregation module to obtain the latent fMRI feature:
\begin{equation}
F_f = \mathcal{FA}(F_{emb}).
\end{equation}
To enhance the information from the class token $T_c^i$, we align it with visual and additional textual feature spaces. Specifically, since we use $N$ frames of fMRI as input, we average the class tokens over all frames:
\begin{equation}
\mathbf{c}_f = \dfrac{1}{N}\sum\limits_i^NT_{c}^{i}.
\end{equation}
Next, to improve the performance of contrastive learning, we use ViT-H CLIP to extract multi-view image and text features. For the multi-view images, we randomly select one of the six rendered views and compute the visual feature using the CLIP vision encoder $E_v$, For text, we directly use the CLIP text encoder to extract the feature:
\begin{equation}
\mathbf{c}_v =E_v(V_k);\ \mathbf{c}_t = E_t(\text{Text}).
\end{equation}
We then calculate the contrastive learning losses between the fMRI features extracted from half of the encoder’s transformer blocks and the visual and textual features, in order to enhance the quality of the extracted features.
\begin{equation}
\mathcal{L}_{fv} = \mathcal{L}_{clip}(\mathbf{c}_f, \mathbf{c}_v);\  \mathcal{L}_{ft} = \mathcal{L}_{clip}(\mathbf{c}_f, \mathbf{c}_t).
\end{equation}
Thus the constrastive loss $\mathcal{L}_c$ for the encoder is defined as:
\begin{equation}
\mathcal{L}_c=\mathcal{L}_{fv} + \mathcal{L}_{ft}.
\end{equation}
During training, we optimize the neuro-align encoder (initialized with pre-trained weights), while keeping the CLIP encoders frozen. It is important to note that the CLIP encoders $E_v$ and $E_t$, along with the images ${\mathbf{V}}$ and text $T$, are used only during training and are discarded during inference.

This approach enables us to align the fMRI features $\mathbf{c}_f$ and embeddings $F_f$ with both the visual and textual spaces. These aligned fMRI features will then serve as the conditional information for the multi-view diffusion model.

\subsection{3D Generation}
\label{3dmodel}
To fully utilize the fMRI signal features and generate textured 3D objects, we design a pipeline that generates multi-view images and synthesizes them into 3D models. Let $\mathbf{c}_f$ represent the fMRI features and $\mathbf{F}_f$ the embeddings derived from the neuro align encoder. The objective is to generate multi-view images $\mathbf{V}^{mv}$ conditioned on these fMRI signals.

In our model, $\mathbf{F}_f$ and $\mathbf{c}_f$ serve as conditional latent inputs for the cross-attention mechanism in the Multi-View Diffusion Model. To improve training efficiency and fully utilize the generation capabilities of the pretrained model, we employ a Low-Rank Adaptation (LoRA)~\cite{lora} fine-tuning strategy. Specifically, LoRA is applied to the projection layers of the query and value matrices in both the cross-attention and self-attention modules, while keeping the rest of the model parameters frozen.

To train the model, we prepare GT multi-view images $\mathbf{V}$ by rendering 3D objects into six target images at a resolution of $512 \times 512$ with a white background in blender. The poses of the six images are defined by interleaving absolute elevations of $20^\circ$ and $-10^\circ$, combined with azimuths relative to the query image, starting at $30^\circ$ and increasing by $60^\circ$ for each subsequent pose. The pretrained diffusion model~\cite{instantmesh} generates $960 \times 640$ images, which represent six multi-view images arranged in a $3 \times 2$ grid. Each of these images is resized to $320 \times 320$ for processing.

During the reverse diffusion process, the Multi-View Diffusion Model $D_{mv}$ estimates the noise $\hat{\mathbf{\epsilon}}_t$ at each timestep $t$, conditioned on the fMRI feature $\mathbf{c}_f$ and embeddings $F_f$:
\begin{equation}
\hat{\mathbf{\epsilon}}_t = D_{mv}(\mathbf{V}_t, t, \mathbf{c}_f, F_f) 
\end{equation}
The training objective minimizes the discrepancy between the predicted noise $\hat{\mathbf{\epsilon}}_t$ and the true noise $\mathbf{\epsilon}$ through the following loss function:
\begin{equation}
\mathcal{L}_{D_{mv}} = \mathbb{E}_{\mathbf{V}, \mathbf{\epsilon}, t} \left[ \| \mathbf{\epsilon} - \hat{\mathbf{\epsilon}}_t \|^2 \right]
\end{equation}
Then, the loss function for our model is $\mathcal{L}$: 
\begin{equation}
\mathcal{L}=\mathcal{L}_c+ \mathcal{L}_{D_{mv}}
\end{equation}
After obtaining the multi-view images $\mathbf{V}_{mv}$ of the 3D object, we employ the off-the-shelf sparse-view LRM~\cite{instantmesh} method to generate the final 3D textured mesh.

\section{Experiments and Benchmark}
To establish new benchmarks for 3D visual decoding, we conduct experiments in both standard and Out-of-Distribution (OOD) settings. In this section, we introduce the metrics and provide details of the experiments.

\begin{table*}
\caption{
\textbf{Performance Comparison on fMRI-3D.} We report the average metrics for each subject, with each subject being trained and tested on their own data, comparing baseline methods and our approaches. LEA-3D and fMRI-PTE-3D are variants of LEA and fMRI-PTE, respectively, and are only compared on fMRI-Shape. MinD-3D serves as the baseline for both fMRI-Shape and fMRI-Objaverse. The metrics on individual subject for MindD-3D++ are also reported, enabling detailed analysis and future comparisons.
\label{tab:metric_table}
}
\centering{
    \setlength{\tabcolsep}{3.2mm}{
    \begin{tabular}{c|c|cc|ccc|ccc}
    \toprule
    \multicolumn{1}{c|}{\multirow{2}{*}{\textsc{Methods}}} & \multicolumn{1}{c|}{\multirow{2}{*}{\textsc{Dataset}}}  & \multicolumn{2}{c|}{Semantic-Level} & \multicolumn{3}{c|}{Structure-Level} & \multicolumn{3}{c}{Textural-Level} \\
       & & 2-way$\uparrow$  &  10-way$\uparrow$   & FPD$\downarrow$ & CD$\downarrow$  & EMD$\downarrow$ & LPIPS$\downarrow$ &PSNR$\uparrow$ & SSIM$\uparrow$ \\
    \bottomrule
    \midrule
    LEA-3D~\cite{qian2023semantic} &\multirow{4}{*}{fMRI-Shape}&  0.787 & 0.371 & 4.229  & 2.291 & 5.347 &0.557 &-& 0.617   \\
    fMRI-PTE-3D~\cite{qian2023fmri}&&  0.815 & 0.392  & 3.571  & 1.992 & 4.621  & 0.462  &-& 0.645   \\
    MinD-3D~\cite{gao2023mind3d}   &&  0.828 & 0.459  & 3.157  & 1.742  & 3.833  &  0.306 &14.98 & 0.674 \\ 
\cellcolor{ForestGreen!8}\textbf{MinD-3D++} &\cellcolor{ForestGreen!8} & \cellcolor{ForestGreen!8} \textbf{0.887} & \cellcolor{ForestGreen!8} \textbf{0.616} & \cellcolor{ForestGreen!8} \textbf{3.025}  & \cellcolor{ForestGreen!8} \textbf{1.635} & \cellcolor{ForestGreen!8} \textbf{3.672}& \cellcolor{ForestGreen!8} \textbf{0.234} & \cellcolor{ForestGreen!8} \textbf{16.44} & \cellcolor{ForestGreen!8} \textbf{0.763}  \\ 
\midrule
\cellcolor{ForestGreen!8} Subject 1  & \cellcolor{ForestGreen!8} & \cellcolor{ForestGreen!8} 0.912 & \cellcolor{ForestGreen!8} 0.645 & \cellcolor{ForestGreen!8} 2.953 & \cellcolor{ForestGreen!8} 1.531 & \cellcolor{ForestGreen!8} 3.419 & \cellcolor{ForestGreen!8} 0.226 & \cellcolor{ForestGreen!8} 17.22 & \cellcolor{ForestGreen!8} 0.783 \\
\cellcolor{ForestGreen!8} Subject 2  & \cellcolor{ForestGreen!8} & \cellcolor{ForestGreen!8} 0.842 & \cellcolor{ForestGreen!8} 0.583 & \cellcolor{ForestGreen!8} 3.078 & \cellcolor{ForestGreen!8} 1.792 & \cellcolor{ForestGreen!8} 3.875 & \cellcolor{ForestGreen!8} 0.242 & \cellcolor{ForestGreen!8} 15.79 & \cellcolor{ForestGreen!8} 0.729 \\
\cellcolor{ForestGreen!8} Subject 3  & \cellcolor{ForestGreen!8} & \cellcolor{ForestGreen!8} 0.869 & \cellcolor{ForestGreen!8} 0.623 & \cellcolor{ForestGreen!8} 3.043 & \cellcolor{ForestGreen!8} 1.642 & \cellcolor{ForestGreen!8} 3.773 & \cellcolor{ForestGreen!8} 0.239 & \cellcolor{ForestGreen!8} 15.98 & \cellcolor{ForestGreen!8} 0.751 \\
\cellcolor{ForestGreen!8} Subject 4  & \cellcolor{ForestGreen!8} & \cellcolor{ForestGreen!8} 0.862 & \cellcolor{ForestGreen!8} 0.598 & \cellcolor{ForestGreen!8} 3.062 & \cellcolor{ForestGreen!8} 1.659 & \cellcolor{ForestGreen!8} 3.795 & \cellcolor{ForestGreen!8} 0.241 & \cellcolor{ForestGreen!8} 15.88 & \cellcolor{ForestGreen!8} 0.737 \\
\cellcolor{ForestGreen!8} Subject 5  & \cellcolor{ForestGreen!8} & \cellcolor{ForestGreen!8} 0.908 & \cellcolor{ForestGreen!8} 0.638 & \cellcolor{ForestGreen!8} 2.985 & \cellcolor{ForestGreen!8} 1.567 & \cellcolor{ForestGreen!8} 3.511 & \cellcolor{ForestGreen!8} 0.228 & \cellcolor{ForestGreen!8} 17.16 & \cellcolor{ForestGreen!8} 0.785 \\
\cellcolor{ForestGreen!8} Subject 6  & \cellcolor{ForestGreen!8} & \cellcolor{ForestGreen!8} 0.907 & \cellcolor{ForestGreen!8} 0.603 & \cellcolor{ForestGreen!8} 3.029 & \cellcolor{ForestGreen!8} 1.626 & \cellcolor{ForestGreen!8} 3.581 & \cellcolor{ForestGreen!8} 0.232 & \cellcolor{ForestGreen!8} 16.65 & \cellcolor{ForestGreen!8} 0.771 \\
\cellcolor{ForestGreen!8} Subject 7  & \cellcolor{ForestGreen!8} & \cellcolor{ForestGreen!8} 0.894 & \cellcolor{ForestGreen!8} 0.613 & \cellcolor{ForestGreen!8} 3.034 & \cellcolor{ForestGreen!8} 1.635 & \cellcolor{ForestGreen!8} 3.672 & \cellcolor{ForestGreen!8} 0.235 & \cellcolor{ForestGreen!8} 16.02 & \cellcolor{ForestGreen!8} 0.768 \\
\cellcolor{ForestGreen!8} Subject 8  & \cellcolor{ForestGreen!8}\multirow{-8}{*}{fMRI-Shape}  & \cellcolor{ForestGreen!8} 0.901 & \cellcolor{ForestGreen!8} 0.627 & \cellcolor{ForestGreen!8} 3.018 & \cellcolor{ForestGreen!8} 1.525 & \cellcolor{ForestGreen!8} 3.498 & \cellcolor{ForestGreen!8} 0.232 & \cellcolor{ForestGreen!8} 16.79 & \cellcolor{ForestGreen!8} 0.778 \\
    \midrule
    MinD-3D~\cite{gao2023mind3d}   & &  0.793  & 0.427 & 4.304  & 2.142 & 5.323 &  0.544 & 9.74 & 0.724 \\ 
    \cellcolor{SkyBlue!10} \textbf{MinD-3D++}  & \cellcolor{SkyBlue!10}\multirow{-2}{*}{fMRI-Objaverse} & \cellcolor{SkyBlue!10} \textbf{0.894}  & \cellcolor{SkyBlue!10}\textbf{0.618}  & \cellcolor{SkyBlue!10}\textbf{3.325}  & \cellcolor{SkyBlue!10}\textbf{1.779} & \cellcolor{SkyBlue!10}\textbf{4.073} &  \cellcolor{SkyBlue!10}\textbf{0.343} & \cellcolor{SkyBlue!10} \textbf{11.57} & \cellcolor{SkyBlue!10}\textbf{0.808} \\ 
    \midrule
\cellcolor{SkyBlue!10} Subject 1  & \cellcolor{SkyBlue!10}  & \cellcolor{SkyBlue!10} 0.909 & \cellcolor{SkyBlue!10} 0.636 & \cellcolor{SkyBlue!10} 3.294 & \cellcolor{SkyBlue!10} 1.698 & \cellcolor{SkyBlue!10} 3.995 & \cellcolor{SkyBlue!10} 0.315 & \cellcolor{SkyBlue!10} 11.96 & \cellcolor{SkyBlue!10} 0.825\\
\cellcolor{SkyBlue!10} Subject 6  & \cellcolor{SkyBlue!10}  & \cellcolor{SkyBlue!10} 0.893 & \cellcolor{SkyBlue!10} 0.616 & \cellcolor{SkyBlue!10} 3.311 & \cellcolor{SkyBlue!10} 1.751 & \cellcolor{SkyBlue!10} 4.032 & \cellcolor{SkyBlue!10} 0.326 & \cellcolor{SkyBlue!10} 11.67 & \cellcolor{SkyBlue!10} 0.811\\
\cellcolor{SkyBlue!10} Subject 7  & \cellcolor{SkyBlue!10}  & \cellcolor{SkyBlue!10} 0.895 & \cellcolor{SkyBlue!10} 0.629 & \cellcolor{SkyBlue!10} 3.319 & \cellcolor{SkyBlue!10} 1.779 & \cellcolor{SkyBlue!10} 4.085 & \cellcolor{SkyBlue!10} 0.347 & \cellcolor{SkyBlue!10} 11.44 & \cellcolor{SkyBlue!10} 0.807\\
\cellcolor{SkyBlue!10} Subject 8  & \cellcolor{SkyBlue!10}  & \cellcolor{SkyBlue!10} 0.882 & \cellcolor{SkyBlue!10} 0.593 & \cellcolor{SkyBlue!10} 3.353 & \cellcolor{SkyBlue!10} 1.863 & \cellcolor{SkyBlue!10} 4.134 & \cellcolor{SkyBlue!10} 0.367 & \cellcolor{SkyBlue!10} 11.20 & \cellcolor{SkyBlue!10} 0.797\\
\cellcolor{SkyBlue!10} Subject 15 & \cellcolor{SkyBlue!10}\multirow{-5}{*}{fMRI-Objaverse} &\cellcolor{SkyBlue!10} 0.891 & \cellcolor{SkyBlue!10} 0.616 & \cellcolor{SkyBlue!10} 3.347 & \cellcolor{SkyBlue!10} 1.802 & \cellcolor{SkyBlue!10} 4.117 & \cellcolor{SkyBlue!10} 0.358 & \cellcolor{SkyBlue!10} 11.23 & \cellcolor{SkyBlue!10} 0.801\\
    \bottomrule
    \end{tabular}}
    }
\end{table*}
    
\subsection{Metrics}
To effectively evaluate the performance of our models in reconstructing 3D objects from fMRI signals, we employ metrics across three primary dimensions: semantic, structural, and texture levels.

\noindent \textbf{Semantic Level.}
To assess the semantic quality of our model, we use standard metrics commonly adopted in previous 2D fMRI studies~\cite{chen2023seeing, chen2023cinematic, ozcelik2022reconstruction, mai2023unibrain, sun2024contrast}, specifically N-way top-K accuracy. We report both 2-way top-1 and 10-way top-1 accuracies, as shown in Tab.~\ref{tab:metric_table}. These metrics are determined by comparing rendered images of the reconstructed objects with the ground truth (GT) images, which include texture.

\noindent \textbf{Structural Level.}
Beyond semantic evaluation, it is crucial to measure how accurately our model captures the geometric structure of objects. We utilize common 3D reconstruction metrics~\cite{liu2022towards, xu2019disn, qian2024pushing}: Fréchet Point Cloud Distance (FPD) (scaled by $\times 10^{-1}$), Chamfer Distance (CD) (scaled by $\times 10^{2}$), and Earth Mover's Distance (EMD) (scaled by $\times 10^{2}$). These metrics are computed by sampling point clouds from both the GT and the generated meshes.

\noindent \textbf{Texture Level.}
To evaluate the quality of the texture and appearance in the reconstructed 3D objects, we use five metrics: Learned Perceptual Image Patch Similarity (LPIPS)~\cite{zhang2018perceptual}, Structural Similarity Index (SSIM), and Peak Signal-to-Noise Ratio (PSNR), all calculated at the RGB pixel level.

All 3D objects are rendered in the same format as the input data for the multi-view diffusion model described in Sec.~\ref{3dmodel}. These metrics are calculated for each frame, and the final scores are obtained by averaging the values across all frames.

\subsection{Implementation Details}
As detailed in Sec.~\ref{sec:3}, each sample in our dataset comprises 8 or 10 fMRI frames. To maximize dataset utilization and apply data augmentation during experiments, we randomly select 6 fMRI frames from each sample for training and use the middle 6 frames for inference. The vision region of interest (ROI), extracted from the original $1023 \times 2514$ 2D fMRI images, is resized to $256 \times 256$ for processing.
During contrastive learning, we use the pretrained ViT-H-14 CLIP vision and text encoders to extract features from images and texts. For the image data, as discussed in Sec.~\ref{3dmodel}, we use Blender to render all 3D objects into six distinct views and randomly select one image, resized to $224 \times 224$, for training. For textual descriptions, we use the category name of each object as the text input for the fMRI-Shape dataset. For the fMRI-Objaverse dataset, we adopt text descriptions sourced from Cap3D\cite{CAP3D}. It is worth mentioning that our contrastive learning is not computed on the class token from the encoder's final transformer layer but rather on results from an intermediate layer.
Regarding the architecture, we configure LoRA with $r=64$ and $\alpha=64$ within the multi-view diffusion model on the Q and V layers in the attention blocks, while keeping other parameters fixed during training. MinD-3D is trained end-to-end in a single stage, with both the encoder and the diffusion model initialized using pre-trained weights. Training each model takes approximately one day on eight A100 GPUs. We evaluate our model on both the fMRI-Shape and fMRI-Objaverse datasets.

\subsection{Experiments on fMRI-Shape}

\subsubsection{Standard Experiment}

As the first effort to model 3D textured imaging within the human brain, we establish a standard experimental setup by training and testing on the pre-split, person-specific Core set in fMRI-Shape. Predefined metrics are used to evaluate the model's performance. Given the task's complexity—which involves multiple brain regions—direct comparisons with existing models are challenging, except with MinD-3D. To reduce training costs, we adapted the LEA and fMRI-PTE models, both trained on the same vision ROIs and demonstrating strong performance. Sharing the same 3D decoder as MinD-3D, these models serve as baselines in our experiments, enabling a more contextual and fair comparison within this novel domain.

In the left part of Fig.~\ref{fig:mind3dplus_vis}, we present qualitative results of MinD-3D++ on the fMRI-Shape dataset. The model consistently generates 3D objects that are structurally and texturally similar to their real counterparts while maintaining semantic integrity in most cases. This underscores the robustness of our approach in handling a challenging task and its ability to produce faithful reconstructions. Importantly, the appearance of the reconstructed objects closely resembles the ground truth, demonstrating the effectiveness of our model.

Tab.~\ref{tab:metric_table} presents averaged metrics at three levels across all subjects alongside the baselines. MinD-3D++ outperforms MinD-3D and the other baselines on both semantic and structural levels, indicating its excellence in generating textured objects with high semantic accuracy and a strong ability to preserve structural similarity. Moreover, MinD-3D++, specifically designed to address the shortcomings of MinD-3D in terms of appearance and texture, significantly outperforms MinD-3D at the textural level.

Together, the qualitative and quantitative results validate the feasibility of reconstructing 3D textured objects from fMRI signals.

\subsubsection{Out-Of-Distribution Experiments}

To effectively utilize a subset of the fMRI-Shape dataset and further assess the generalization capabilities of our proposed MinD-3D++ model, we conduct two Out-Of-Distribution (OOD) experiments under challenging settings:
\textbf{1) Across-Person (AP) Testing:} In AP testing, 
we evaluate our model, which was trained only on Subject 1, using the data from Subject 9. We compare the results with the baselines and report the metrics in Tab.~\ref{tab:AP_APAC}.
\textbf{2) Across-Person \& Across-Class (APAC) Testing:} In APAC testing, 
we similarly evaluate our model, trained solely on Subject 1, with the data from Subject 11. We also compare with the baselines and report the metrics in Tab.~\ref{tab:AP_APAC}.

\begin{figure}
\centering
\includegraphics[width=\linewidth]{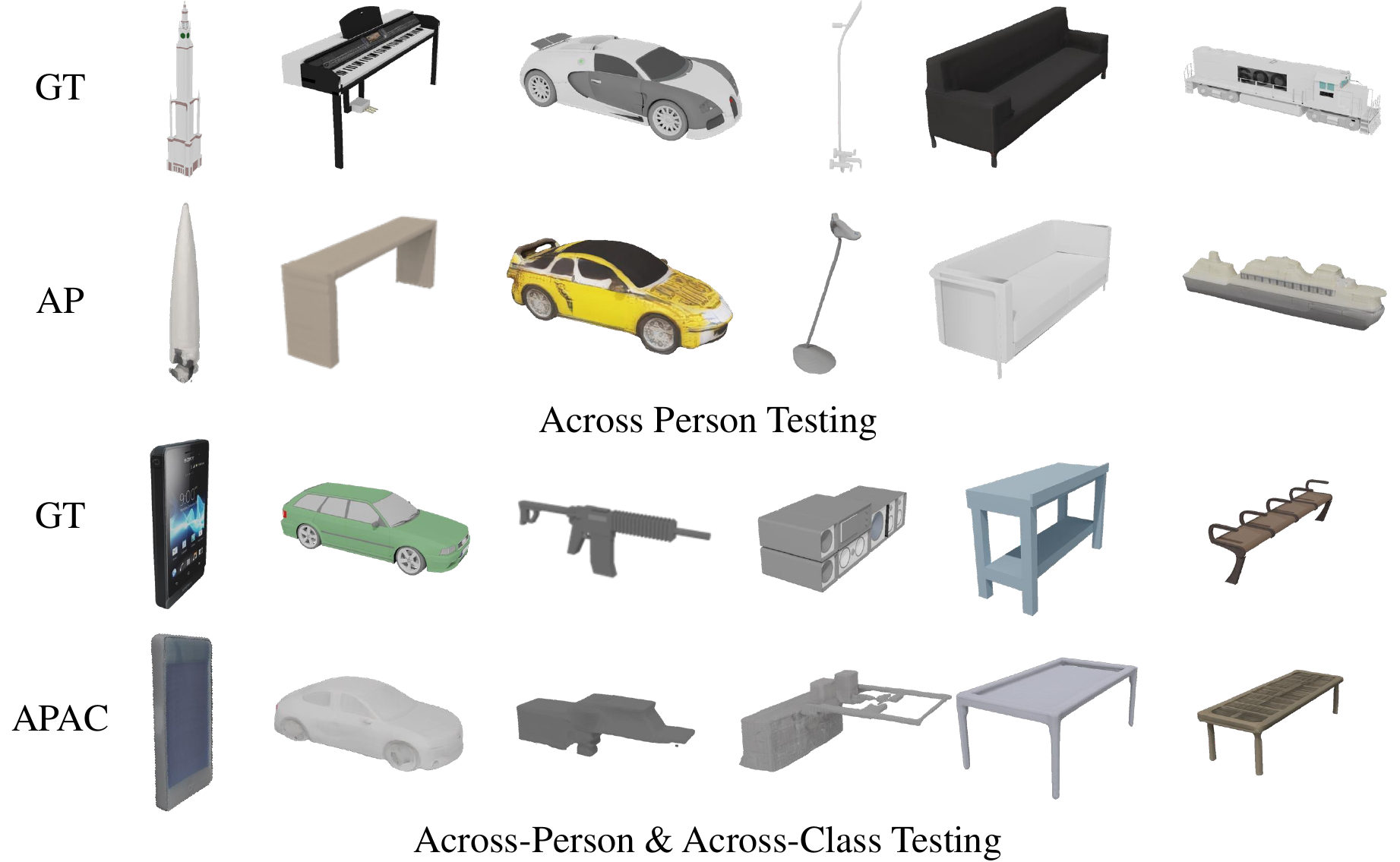}
\caption{
\textbf{Visualization of AP \& APAC testing.} AP Testing trains on Subject 1 and tests on Subject 9. APAC Testing trains on Subject 1 and tests on Subject 11.
\label{fig:vis_apac_more}
}
\end{figure}

We present the reconstructed objects from the AP and APAC tests in Fig.~\ref{fig:vis_apac_more}. As shown in Fig.~\ref{fig:fmri}, individual differences significantly impact the results, as confirmed by our empirical AP and APAC tests. Despite the high difficulty, MinD-3D++ successfully recovers the basic shapes of the objects, establishing a strong baseline for the community.
We include the complete table of all metrics and subjects for the AP and APAC tests in the supplementary material. In Tab.~8, we compare the AP results with those from the regular experiment.
While performance in these out-of-distribution (OOD) scenarios does not reach in-distribution (ID) levels—an expected outcome due to the task’s complexity and substantial individual differences and domain gaps—our method still outperforms existing baselines. This demonstrates the robustness of MinD-3D++ and establishes a new benchmark for future work. 

\begin{table}
  \caption{Quantitative results of AP and APAC testings. We use the model trained on Subject 1 and compare metrics for Subjects 9 and 11 separately.}
  \label{tab:AP_APAC}
  \centering
    {
        \begin{tabular}{l|ccc|ccc}
        \toprule
         \multicolumn{1}{c|}{\multirow{2}{*}{\textsc{Methods}}} & \multicolumn{3}{c|}{AP} & \multicolumn{3}{c}{APAC} \\
         & FPD$\downarrow$ & CD$\downarrow$  & EMD$\downarrow$  & FPD$\downarrow$ & CD$\downarrow$  & EMD$\downarrow$  \\
        \midrule
        \midrule
            LEA-3D          & 5.362   & 3.627  & 6.174  & 6.958  & 4.944  & 8.107  \\
            fMRI-PTE-3D     & 4.501   & 2.956  & 5.772  & 6.261  & 4.570  & 7.843  \\
            MinD-3D         & 3.838   & 2.415  & 5.117  & 5.689  & 4.181  & 7.194   \\ 
    \textbf{MinD-3D++}  & \textbf{3.582} & \textbf{2.237}   & \textbf{4.975}  & \textbf{5.347} & \textbf{3.912} & \textbf{6.915}  \\
        \bottomrule
        \end{tabular}
    }
\end{table}

\begin{figure}
    \centering
    \includegraphics[width=\linewidth]{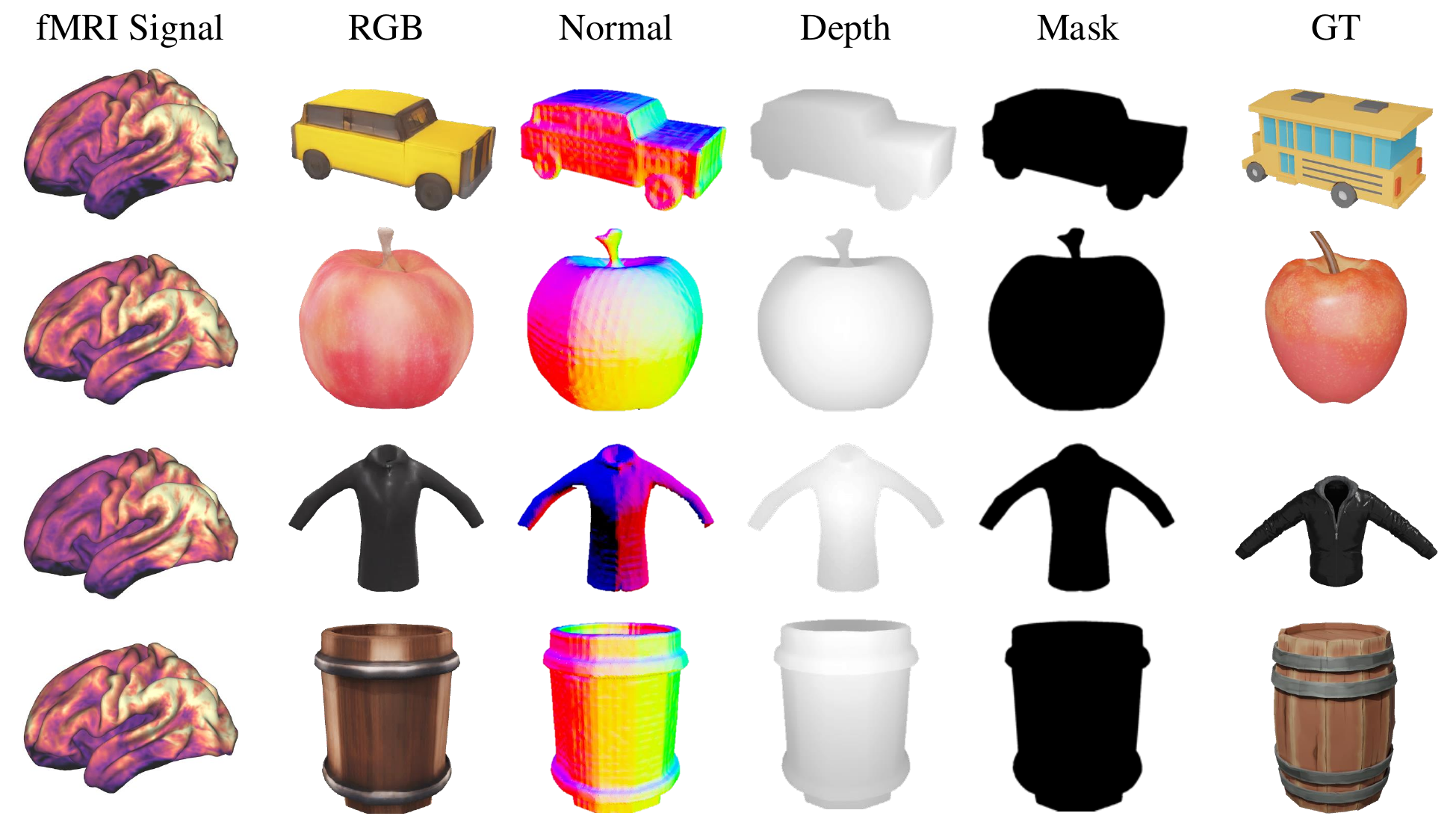}
\caption{
\textbf{Visualization of 3D Details.} We visualize the 3D textured reconstruction process from fMRI signals, using RGB images, surface normals, depth maps, masks, and ground truth (GT) from fMRI-Objaverse.
\label{fig:3d_vis}
}
\end{figure}

\subsection{Experiments on fMRI-Objaverse}
The fMRI-Objaverse dataset presents a more challenging and larger-scale environment, increasing the difficulty of effective modeling. For this experiment, we randomly partitioned the objects into 2,709 for training and 432 for testing. Our goal was to reconstruct textured 3D objects from human brain data and rigorously evaluate our models' capabilities. To this end, we trained MinD-3D++ on the pre-split fMRI-Objaverse dataset, using MinD-3D as our baseline for comparison.

As detailed in the lower section of Tab.~\ref{tab:metric_table}, we report metrics at three levels for our proposed model and all the baselines on the fMRI-Objaverse dataset. Across all metrics, MinD-3D++ outperforms MinD-3D and other baselines. Notably, on the more complex fMRI-Objaverse dataset—compared to the simpler fMRI-Shape—the performance improvement of MinD-3D++ is markedly greater, underscoring its robustness and enhanced capacity.

The right side of Fig.~\ref{fig:mind3dplus_vis} presents qualitative results of MinD-3D++ on the fMRI-Objaverse dataset. Beyond capturing structural aspects comparable to MinD-3D, MinD-3D++ excels at accurately reconstructing appearance and color details.
Furthermore, Fig.~\ref{fig:3d_vis} offers additional 3D details. This figure displays the fMRI signals alongside corresponding RGB images, normals, depths, and masks, compared with ground truth (GT), to showcase the quality of the reconstructions results of MinD-3D++.

Given the substantially larger scale of the fMRI-Objaverse dataset relative to fMRI-Shape, the full capacity of MinD-3D++ is effectively leveraged. The high-quality textured 3D mesh reconstructions obtained from fMRI signals provide compelling evidence for the feasibility of reconstructing detailed textured 3D objects based on human brain data.

\subsection{Ablation Study}

To demonstrate the effectiveness of our proposed MinD-3D++, we conduct an ablation study on the contrastive loss. Specifically, we perform experiments in which we remove the contrastive loss, and separately remove the image and text branches from the contrastive loss, computing the loss based on the final transformer's output from the encoder. We then report 2-way, 10-way, LPIPS, PSNR, and SSIM metrics compared with the full model in Tab.~\ref{tab:Ablation_contrastive}, which validates the effectiveness of incorporating image and text information and introducing contrastive learning at intermediate transformer layers.

\begin{table}
  \caption{Ablation study of contrastive learning in MinD-3D++. Metrics on the fMRI-Shape dataset compare MinD-3D++ with baseline models.}
  \label{tab:Ablation_contrastive}
  \centering
    {
        \begin{tabular}{l|ccccc}
        \toprule
         \multicolumn{1}{c|}{\multirow{1}{*}{\textsc{Methods}}} & 2-way$\uparrow$ & 10-way$\uparrow$  & LPIPS$\downarrow$ & PSNR$\uparrow$ & SSIM$\uparrow$  \\
        \midrule
        \midrule
            w/o  contrastive   & 0.830   & 0.463  & 0.296  & 15.17 & 0.697   \\
            w/o  image         & 0.843   & 0.549  & 0.266  & 15.79 & 0.712   \\
            w/o  text          & 0.858   & 0.563  & 0.252  & 15.93 & 0.734   \\
            Add on 12th        & 0.875   & 0.604  & 0.248  & 16.24 & 0.759   \\
            \midrule
    \textbf{Full model}  & \textbf{0.887} & \textbf{0.616}   & \textbf{0.234}  & \textbf{16.44} & \textbf{0.763}   \\ 
        \bottomrule
        \end{tabular}
    }
\end{table}

\section{Analysis on Dataset}

\subsection{Analysis for Object Angle Variations}\label{sec:7.1}

In this section, we explore brain activity patterns associated with object angle changes. Specifically, given whole-brain fMRI data collected while viewing 3D objects from different angles, we perform an angle classification task using simple linear classification on each subject in the fMRI-Shape dataset individually. We construct the fMRI data using views of objects from two different angles (the 3rd and 9th frames of each trial), drawn from both training and testing sets. We then randomly select 70\% of the data for training and use the remaining 30\% for testing. The average classification accuracy across all subjects for this angle-specific task reaches 88.10\%.

To evaluate voxel importance, we visualize the absolute values of the classifier weights for Subject 1 in Fig.~\ref{fig:angle_vis}. Generally, the weight assigned to each voxel reflects its importance, with red regions indicating voxels with higher coefficients and other regions showing lower values. To better understand the relationship between ROIs and angle discrimination, we compute the mean value of the classifier weights within each HCP-MMP-defined ROI to quantify its importance. We then list the top-8 ROIs for each subject, along with the overall top-8 ROIs, calculated by summing the weights of each ROI across all subjects in the fMRI-Shape dataset, as shown in Tab.~\ref{tab:angle_roi}. We also report the classification accuracy for each subject and the overall performance. Notably, ROIs such as V8, V3A, and V4 consistently appear among the top regions across all eight subjects, suggesting their common role in distinguishing object angles.

Furthermore, we conduct paired \emph{t}-tests to evaluate whether these three ROIs are significantly more important than the primary visual area V1. The resulting \emph{p}-values are $1.38 \times 10^{-5}$, $2.15 \times 10^{-2}$, and $1.24 \times 10^{-5}$, respectively. All values fall below the significance threshold of $5 \times 10^{-2}$, indicating that these ROIs are significantly more important than V1 in the context of angle discrimination.

\subsection{Analysis of Different Objects}
In this section, we explore the brain activity patterns associated with viewing different objects. The analysis is divided into two parts:

\begin{figure}
\centering
\includegraphics[width=\linewidth]{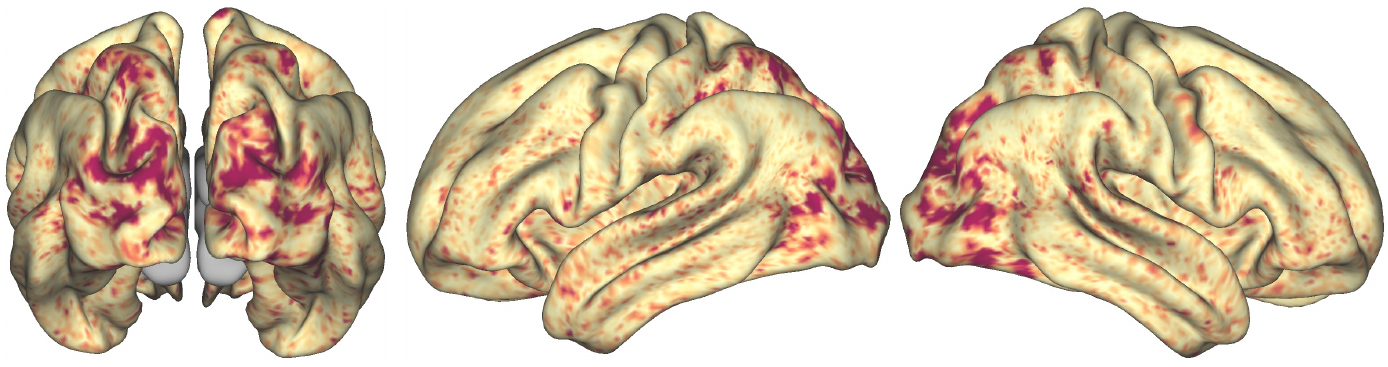}
\caption{
\textbf{Visualization of Weights for Object Angles.} The absolute linear classifier weights are shown for voxels across the whole brain to assess the importance of each ROI in distinguishing objects at different angles. Red regions indicate voxels with higher coefficients.
\label{fig:angle_vis}
}
\end{figure}

\begin{table}
\caption{\textbf{ROI Importance for Angle Classification.} We report the top-8 ROIs for object angle discrimination for each subject. Consistently selected ROIs such as V8, V3A, and V4 suggest importance in encoding angle variations. Classification accuracy for each subject is also provided.}

  \label{tab:angle_roi}
  \centering
    \setlength{\tabcolsep}{1.5mm}{
    \begin{tabular}{c|c|l}
    \toprule
     \multicolumn{1}{c|}{\multirow{1}{*}{\textsc{Subject}}} & Accuracy & Top 8 Important ROIs \\
    \midrule
    \midrule
        1   & 91.34  & \setlength{\fboxsep}{0pt}\colorbox{purple!15}{\strut V3A}, MT, V7, \setlength{\fboxsep}{0pt}\colorbox{cyan!15}{\strut V8}, V3, V3CD, \setlength{\fboxsep}{0pt}\colorbox{orange!15}{\strut V4}, V3B \\
        2   & 84.88  & \setlength{\fboxsep}{0pt}\colorbox{cyan!15}{\strut V8}, \setlength{\fboxsep}{0pt}\colorbox{orange!15}{\strut V4}, V3, \setlength{\fboxsep}{0pt}\colorbox{purple!15}{\strut V3A}, VMV3, PHA1, MST, V1 \\
        3   & 88.82  & VMV3, \setlength{\fboxsep}{0pt}\colorbox{cyan!15}{\strut V8}, PIT, \setlength{\fboxsep}{0pt}\colorbox{purple!15}{\strut V3A}, PEF, VVC, \setlength{\fboxsep}{0pt}\colorbox{orange!15}{\strut V4}, FFC \\
        4   & 92.42  & MST, \setlength{\fboxsep}{0pt}\colorbox{purple!15}{\strut V3A}, LO1, \setlength{\fboxsep}{0pt}\colorbox{cyan!15}{\strut V8}, V7, IPS1, \setlength{\fboxsep}{0pt}\colorbox{orange!15}{\strut V4}, V4t \\
        5   & 83.07  & \setlength{\fboxsep}{0pt}\colorbox{cyan!15}{\strut V8}, \setlength{\fboxsep}{0pt}\colorbox{orange!15}{\strut V4}, LO1, V3CD, \setlength{\fboxsep}{0pt}\colorbox{purple!15}{\strut V3A}, V3, VMV3, V2 \\
        6   & 85.26  & \setlength{\fboxsep}{0pt}\colorbox{cyan!15}{\strut V8}, MST, VVC, MT, \setlength{\fboxsep}{0pt}\colorbox{cyan!15}{\strut V4}, FFC, 25, V3B \\
        7   & 92.40  & \setlength{\fboxsep}{0pt}\colorbox{purple!15}{\strut V3A}, MST, LO1, \setlength{\fboxsep}{0pt}\colorbox{cyan!15}{\strut V8}, V7, V4t, \setlength{\fboxsep}{0pt}\colorbox{orange!15}{\strut V4}, IPS1 \\
        8   & 86.59  & FEF, V6, VMV1, \setlength{\fboxsep}{0pt}\colorbox{cyan!15}{\strut V8}, VMV2, VVC, VMV3, 6v \\
    \midrule
    Average  & 88.10 & \setlength{\fboxsep}{0pt}\colorbox{cyan!15}{\strut V8}, \setlength{\fboxsep}{0pt}\colorbox{purple!15}{\strut V3A}, \setlength{\fboxsep}{0pt}\colorbox{orange!15}{\strut V4}, MST, VMV3, LO1, V3, V7 \\
    \bottomrule
    \end{tabular}
    }
\end{table}

\begin{figure}
\centering
\includegraphics[width=\linewidth]{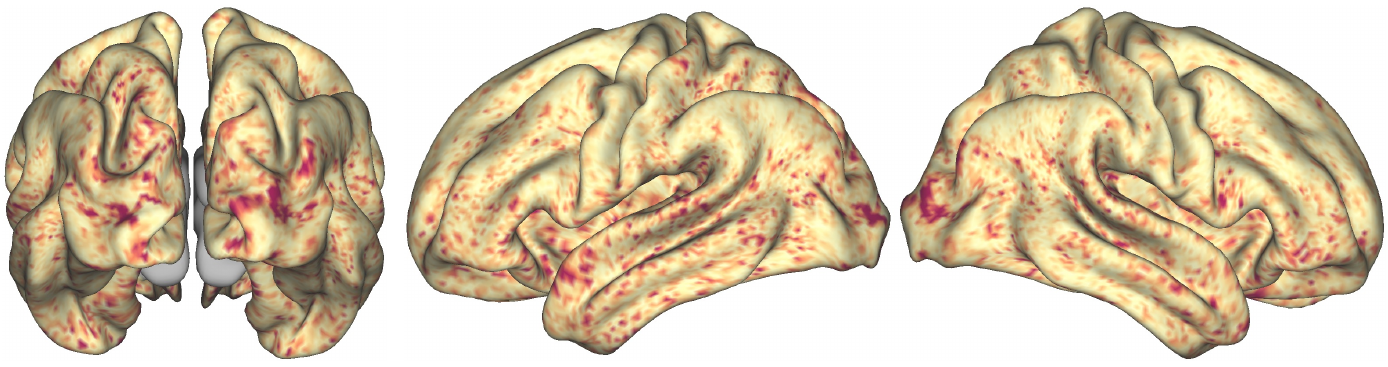}
\caption{
\textbf{Visualization of Weights for Object Types.} 
The absolute linear classifier weights are shown for voxels across the brain to explore the importance of each ROI in classifying different object types (cars and rifles). Red regions indicate voxels with higher coefficients.
\label{fig:cls_vis}
}
\end{figure}

\begin{table}
\caption{\textbf{ROI Importance for Object Classification.} We report the top-8 ROIs associated with object angle discrimination per subject. Consistently selected ROIs such as LO1, s32, and 25 highlight significance in object recognition. Classification accuracy for each subject is also provided.}
  \label{tab:object_cls_roi}
  \centering
  {
    \begin{tabular}{c|c|l}
    \toprule
     \multicolumn{1}{c|}{\multirow{1}{*}{\textsc{Subject}}} & Accuracy & Top 8 Important ROIs \\
    \midrule
    \midrule
        1 & 48.75 & \setlength{\fboxsep}{0pt}\colorbox{cyan!15}{\strut LO1}, PGp, V7, V3CD, V3, FFC, V4, V4t \\
        2 & 54.67 & PGp, V3CD, \setlength{\fboxsep}{0pt}\colorbox{purple!15}{\strut s32}, \setlength{\fboxsep}{0pt}\colorbox{orange!15}{\strut 25}, STGa, OFC, TA2, \setlength{\fboxsep}{0pt}\colorbox{cyan!15}{\strut LO1} \\
        3 & 43.16 & \setlength{\fboxsep}{0pt}\colorbox{cyan!15}{\strut LO1}, FST, V4, V4t, \setlength{\fboxsep}{0pt}\colorbox{purple!15}{\strut s32}, FFC, V6A, V3B \\
        4 & 48.22 & V8, PIT, V3, \setlength{\fboxsep}{0pt}\colorbox{purple!15}{\strut s32}, V7, \setlength{\fboxsep}{0pt}\colorbox{orange!15}{\strut 25}, FFC, V2 \\
        5 & 51.96 & \setlength{\fboxsep}{0pt}\colorbox{cyan!15}{\strut LO1}, \setlength{\fboxsep}{0pt}\colorbox{orange!15}{\strut 25}, \setlength{\fboxsep}{0pt}\colorbox{purple!15}{\strut s32}, V7, a24, V3CD, LO3, PGp \\
        6 & 43.54 & \setlength{\fboxsep}{0pt}\colorbox{purple!15}{\strut s32}, MT, 10r, OFC, \setlength{\fboxsep}{0pt}\colorbox{cyan!15}{\strut LO1}, LO3, \setlength{\fboxsep}{0pt}\colorbox{orange!15}{\strut 25}, V4 \\
        7 & 48.43 & MT, V4t, FST, V3, \setlength{\fboxsep}{0pt}\colorbox{cyan!15}{\strut LO1}, PH, V4, MST \\
        8 & 45.76 & \setlength{\fboxsep}{0pt}\colorbox{cyan!15}{\strut LO1}, \setlength{\fboxsep}{0pt}\colorbox{orange!15}{\strut 25}, \setlength{\fboxsep}{0pt}\colorbox{purple!15}{\strut s32}, V4, 52, STGa, V3CD, a24 \\
    \midrule
    Average & 48.10 & \setlength{\fboxsep}{0pt}\colorbox{cyan!15}{\strut LO1}, \setlength{\fboxsep}{0pt}\colorbox{purple!15}{\strut s32}, \setlength{\fboxsep}{0pt}\colorbox{orange!15}{\strut 25}, V4, PGp, V3, V7, V3CD \\
    \bottomrule
    \end{tabular}
  }
\end{table}

\subsubsection{Between Different Types of Objects}

To observe and analyze how the brain responds to different object categories, we also perform a 13-class classification task by applying logistic regression to the whole-brain fMRI signals per subject in fMRI-Shape. The averaged accuracy of all the subjects achieved is 48.10\%. Detailed classification accuracies for all subjects in the fMRI-Shape dataset, along with the top eight most important ROIs, are summarized in Tab.~\ref{tab:object_cls_roi}. The calculation of the top ROIs follows the same approach used for object angle variations (see Sec.~\ref{sec:7.1}).

We observe that ROIs such as LO1, s32, and area 25 consistently appear among the top regions across all eight subjects, suggesting their shared involvement in object recognition. Similarly, we conduct paired \emph{t}-tests to evaluate whether these three ROIs are significantly more important than the primary visual area V1. The resulting p-values are $1.10 \times 10^{-2}$, $8.18 \times 10^{-4}$, and $1.22 \times 10^{-2}$, respectively. The clear gap between these p-values and the significance threshold of $5 \times 10^{-2}$ further demonstrates the greater importance of these ROIs compared to V1 in the context of object recognition.
Visualization of the absolute values of the classifier weights is given in Fig.~\ref{fig:cls_vis}, taking Subject 1 as an example. In this figure, red regions indicate voxels with higher coefficients; other colors show lower values.

\begin{figure}
\centering
\includegraphics[width=\linewidth]{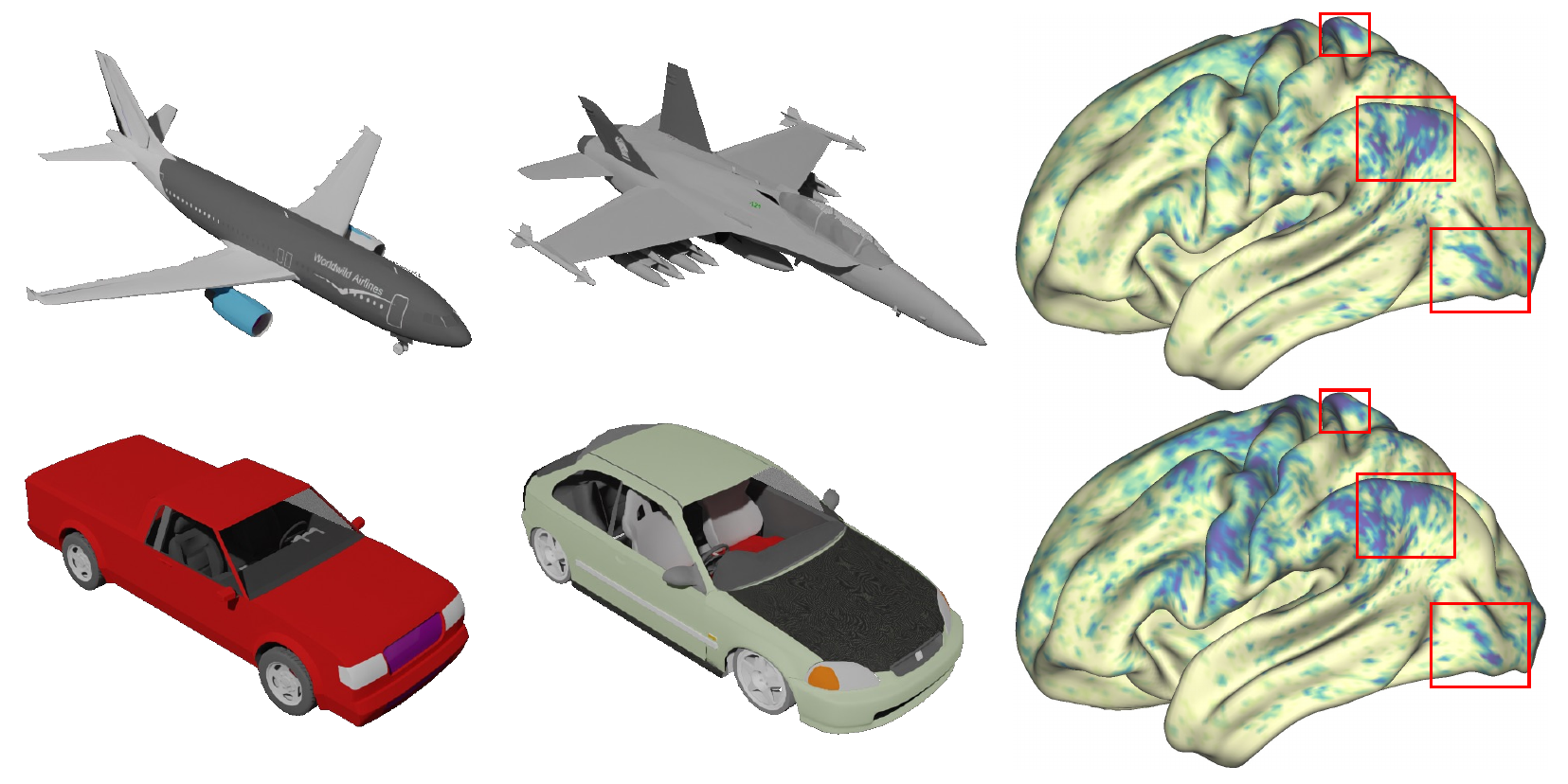}
\caption{
\textbf{Differentiation between objects within the same category.} We show the differentiation between two cars and two planes to illustrate how the brain distinguishes objects within the same category in Subject 1. Deep blue indicates voxels with higher values.
\label{fig:diff_obj}
}
\end{figure}

\begin{table}
  \caption{\textbf{Top 8 ROIs for Intra-class Classification in Subject 1.} We report the top 8 ROIs with the highest importance, showing that the same subject tends to engage overlapping brain regions when distinguishing intra-class objects.}
  \label{tab:intra_cls_roi_subject1}
  \centering
    \setlength{\tabcolsep}{3.5mm}{
    \begin{tabular}{c|l}
    \toprule
    \multicolumn{1}{c|}{\textsc{Category}} & Top 8 Important ROIs \\
    \midrule
    \midrule
Bench & PEF, LO2, 3b, 6v, FOP2, 6d, PHA2, 47m \\
Cabinet & LO2, MST, IP1, \setlength{\fboxsep}{0pt}\colorbox{purple!15}{\strut 7AL}, V4, TPOJ2, V3, TA2 \\
Car & 5L, \setlength{\fboxsep}{0pt}\colorbox{cyan!15}{\strut PFm}, 6v, VMV3, FOP2, PF, \setlength{\fboxsep}{0pt}\colorbox{purple!15}{\strut 7AL}, 4 \\
Chair & \setlength{\fboxsep}{0pt}\colorbox{purple!15}{\strut 7AL}, \setlength{\fboxsep}{0pt}\colorbox{orange!15}{\strut 7Am}, \setlength{\fboxsep}{0pt}\colorbox{cyan!15}{\strut PFm}, 5L, 7PC, VIP, PGs, PGi \\
Display & \setlength{\fboxsep}{0pt}\colorbox{cyan!15}{\strut PFm}, V4t, PGs, V3CD, LO2, VIP, 8Av, p9-46v \\
Lamp & \setlength{\fboxsep}{0pt}\colorbox{orange!15}{\strut 7Am}, \setlength{\fboxsep}{0pt}\colorbox{purple!15}{\strut 7AL}, \setlength{\fboxsep}{0pt}\colorbox{cyan!15}{\strut PFm}, 7PC, 6ma, VIP, 6d, 1 \\
Phone & \setlength{\fboxsep}{0pt}\colorbox{purple!15}{\strut 7AL}, \setlength{\fboxsep}{0pt}\colorbox{cyan!15}{\strut PFm}, \setlength{\fboxsep}{0pt}\colorbox{orange!15}{\strut 7Am}, LO2, 6mp, IP2, 1, 5L \\
Plane & V4t, 6v, 6ma, 6d, 6a, 10d, 9p, \setlength{\fboxsep}{0pt}\colorbox{cyan!15}{\strut PFm} \\
Rifle & PHA2, OP2-3, 10d, MST, TA2, V3B, OP4, PI \\
Sofa & \setlength{\fboxsep}{0pt}\colorbox{cyan!15}{\strut PFm}, PGs, \setlength{\fboxsep}{0pt}\colorbox{purple!15}{\strut 7AL}, \setlength{\fboxsep}{0pt}\colorbox{orange!15}{\strut 7Am}, IP2, VIP, 5L, PEF \\
Speaker & LO2, TPOJ2, V3B, V3A, 10d, V3, MST, TA2 \\
Table & STSda, OP4, 43, FOP2, TA2, Ig, V6A, STSva \\
Vessel & \setlength{\fboxsep}{0pt}\colorbox{cyan!15}{\strut PFm}, \setlength{\fboxsep}{0pt}\colorbox{purple!15}{\strut 7AL}, \setlength{\fboxsep}{0pt}\colorbox{orange!15}{\strut 7Am}, IP2, 6d, 7PC, 6mp, 6ma \\
    \midrule
    \bottomrule
    \end{tabular}
    }
\end{table}

\subsubsection{Between Objects Within the Same Category}

In this case, classification experiments are more challenging due to the lack of specific labels for supervision. Therefore, we first explore the ROIs involved in recognizing intra-class objects for the same subject. As shown in Fig.~\ref{fig:diff_obj}, we visualize the absolute voxel-wise difference between two objects (a car and a plane) for analysis. In the figure, deep blue indicates voxels with higher values, while lighter colors represent lower values. Red rectangles highlight regions with particularly high values. This suggests that the ROIs required for recognizing intra-class objects within the same subject are correlated to some extent.

To further investigate this, we compute the mean absolute difference across all pairwise combinations of fMRI signals within the same object category, as shown in the following equation, to obtain the weight for each voxel:
\begin{equation}
W_{\text{intra}} = \frac{2}{n(n - 1)} \sum_{i < j} \left| f_i - f_j \right|.
\end{equation}
We then average the voxel-wise weights within each ROI defined by the HCP-MMP atlas to quantify the importance of each region and select the top 8 ROIs based on their scores. We first take Subject 1 as an example to analyze the key ROIs involved in recognizing different intra-class objects. As shown in Tab.~\ref{tab:intra_cls_roi_subject1}, we present the top 8 ROIs for Subject 1 when classifying intra-class objects across 13 different object categories in fMRI-Shape. Among them, regions such as PFm, 7AL, and 7Am consistently contribute to the recognition of most objects. 
The remaining top 8 ROIs for other subjects are in Tab.~1 to Tab.~7 in the supplementary material. 
To further examine whether consistent ROIs exist across subjects when recognizing intra-class objects, we compute the average weights across 13 categories for each subject using the same method applied to Subject 1. The top 8 ROIs for each subject across all categories are reported in Tab.~\ref{tab:intra_cls_roi_all_subjects}. We do not observe clear overlaps in these ROIs across subjects, which is understandable given the individual cognitive variability shown in Fig.~\ref{fig:fmri}.

\begin{table}
  \caption{\textbf{Top 8 ROIs for Intra-class Classification Across Subjects.} We report the top 8 ROIs with the highest importance across all object classes in fMRI-Shape for each subject, showing that due to individual cognitive differences, different subjects tend to rely on different ROIs when distinguishing intra-class objects.}
  \label{tab:intra_cls_roi_all_subjects}
  \centering
    \setlength{\tabcolsep}{4mm}{
    \begin{tabular}{c|l}
    \toprule
    \multicolumn{1}{c|}{\textsc{Subject}} & Top 8 Important ROIs \\
    \midrule
    \midrule
    1   & PFm, 7AL, 7Am, LO2, PGs, IP2, 1, 7PC \\
    2   & 5L, 7AL, 7PL, VIP, 5m, 6ma, 7Am, 1 \\
    3   & PGs, 9a, 10d, PGp, p10p, 7PL, V6A, VIP \\
    4   & 9a, 9p, 5L, 8BL, 10d, V6A, 9m, LO2 \\
    5   & 5L, 10d, 7AL, 6v, 9a, 8BL, V6A, 5m \\
    6   & PF, IP1, PFt, PFm, IP2, 1, 7PC, PFop \\
    7   & 9a, 9p, 5L, 8BL, 10d, V6A, PFm, LO2 \\
    8   & V3CD, LO1, V3A, MT, LO2, V7, V3, PGp \\
    \midrule
    \bottomrule
    \end{tabular}
    }
\end{table}

\begin{figure}
\centering
\includegraphics[width=\linewidth]{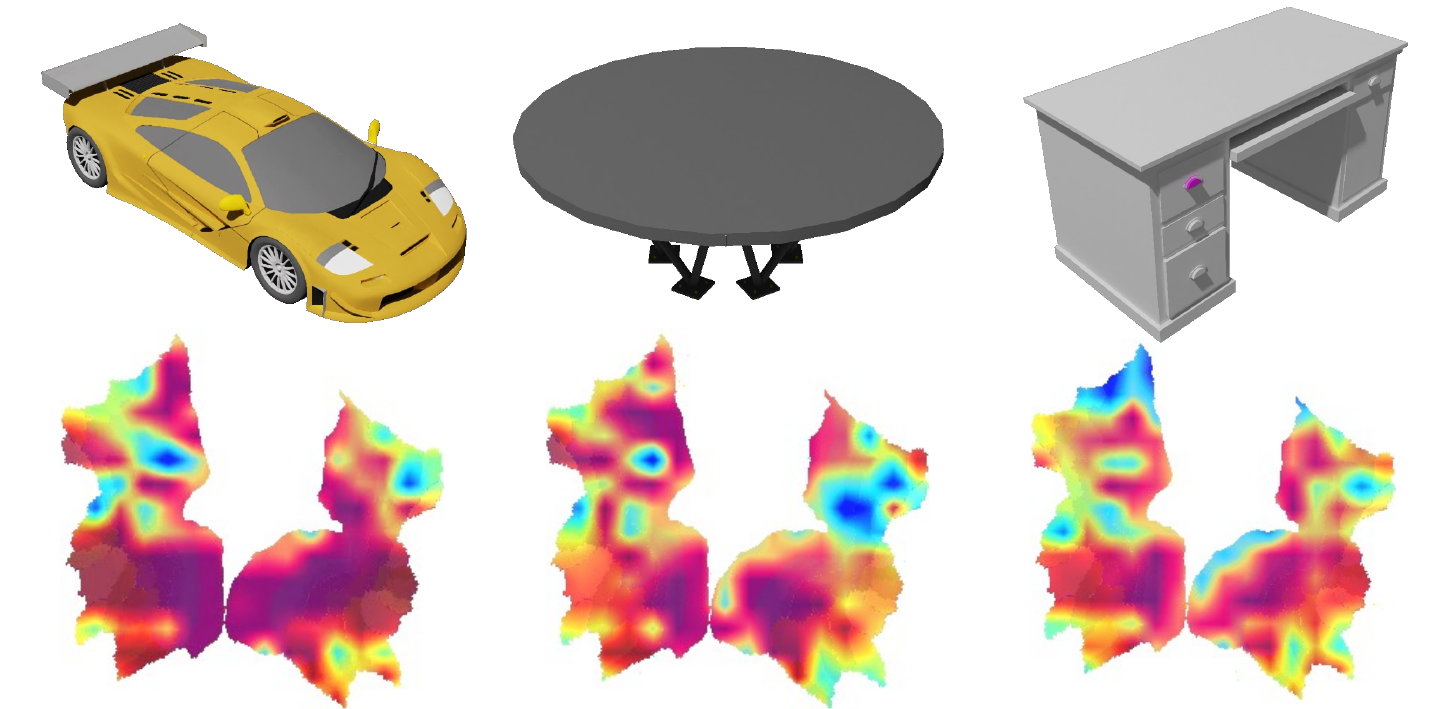}
\caption{
\textbf{CAM for Different Objects.} We use CAM to visualize the importance of each ROI in the visual regions for different objects.
\label{fig:cam_vis}
}
\end{figure}

\subsection{Explore how our brain understands semantics}

MinD-3D++ successfully reconstructs 3D objects both semantically and structurally, so we aim to explore how the brain processes their semantic information. To this end, we use Class Activation Mapping (CAM)\cite{zhou2016cvpr} to analyze the importance of each part of the input fMRI frame. CAMs for three objects are shown in Fig.\ref{fig:cam_vis}. From these, we identify the following ROIs potentially involved in visual semantic processing: FFC, FST, IP0, MT, MST, PHA1, PHA2, PHA3, PH, PIT, V1, V2, V3, V3A, V4, V4t, V6A, V7, V8, VMV1, VMV2, VMV3, and VVC.

\section{Conclusion}

In this paper, we introduce the innovative task of Recon3DMind with texture, alongside the first large-scale dataset—fMRI-3D—across various settings. Technologically, we present a novel end-to-end framework, MinD-3D++, which integrates multiple brain regions, including those associated with human 3D vision, specifically designed for this task. This approach not only establishes new benchmarks in the field but also demonstrates the feasibility of reconstructing textured 3D objects from human brain data.
Our model begins by proficiently extracting features from fMRI signals using a contrastive learning loss. It then generates multi-view images of target objects, which are subsequently used to reconstruct textured 3D models. Comprehensive experimental results and analyses confirm the effectiveness of MinD-3D++ in accurately extracting fMRI features and converting them into their corresponding 3D objects.
Additionally, we perform an in-depth analysis of our proposed fMRI-3D dataset and examine the features extracted by MinD-3D. These evaluations further validate both the quality of the fMRI-3D dataset and the effectiveness of our approach. This pioneering work not only opens a new avenue in neuroimaging and 3D reconstruction but also paves the way for future research aimed at a deeper understanding and visualization of neural representations in 3D vision.







\ifCLASSOPTIONcaptionsoff
  \newpage
\fi



%
\bibliographystyle{IEEEtran}
\bibliography{egbib}

%

\begin{IEEEbiography}[{\includegraphics[width=1in,height=1.25in,clip,keepaspectratio]{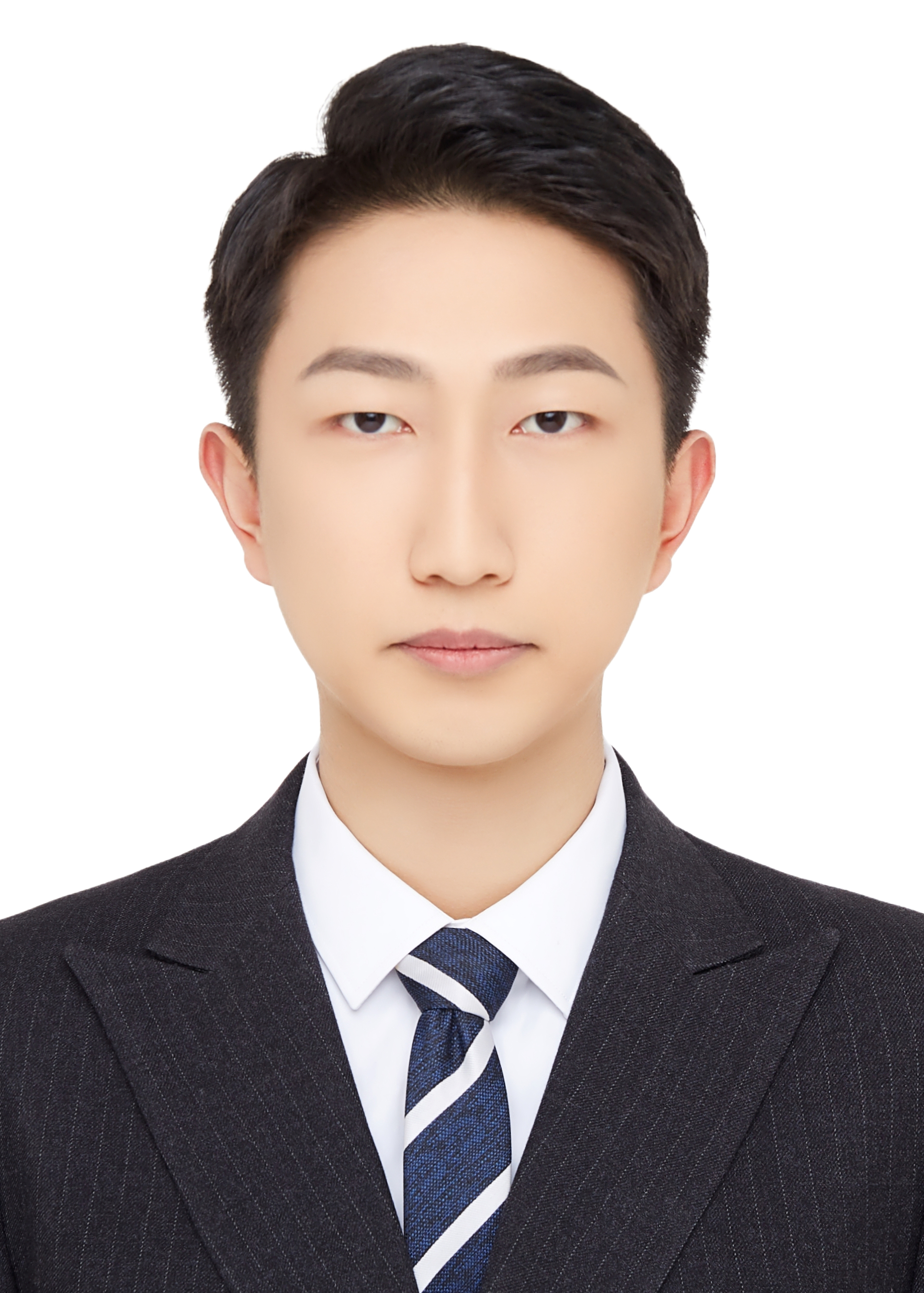}}]{Jianxiong Gao} received the B.S. degree in Statistics from Shandong University in 2022. 
He is currently pursuing a Ph.D. degree in Biomedical Engineering at the Institute of Science and Technology for Brain-Inspired Intelligence, Fudan University, under the supervision of Dr. Yanwei Fu and Dr. Jianfeng Feng. His research interests include video generation and neural decoding.
\end{IEEEbiography}

\begin{IEEEbiography}[{\includegraphics[width=1in,height=1.25in,clip,keepaspectratio]{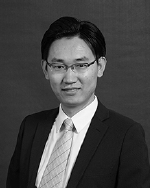}}]{Yanwei Fu} received the MEng degree from the Department of Computer Science and Technology, Nanjing University, China, in 2011, and the PhD degree from the Queen Mary University of London, in 2014. He held a post-doctoral position at Disney Research, Pittsburgh, PA, from 2015 to 2016. He is currently a tenure-track professor at Fudan University.  
He was appointed as the Professor of Special Appointment
(Eastern Scholar) at Shanghai Institutions
of Higher Learning in 2017, and awarded
the 1000 Young talent scholar in 2018. 
His work has led to many awards, including the IEEE ICME 2019 best paper.
He published more than 110 journal/conference papers including IEEE TPAMI, TMM, ECCV, and CVPR. His research interests are one-shot learning,  learning-based 3D reconstruction, and learning-based robotic grasping.
\end{IEEEbiography}

\begin{IEEEbiography}[{\includegraphics[width=1in,height=1.25in,clip,keepaspectratio]{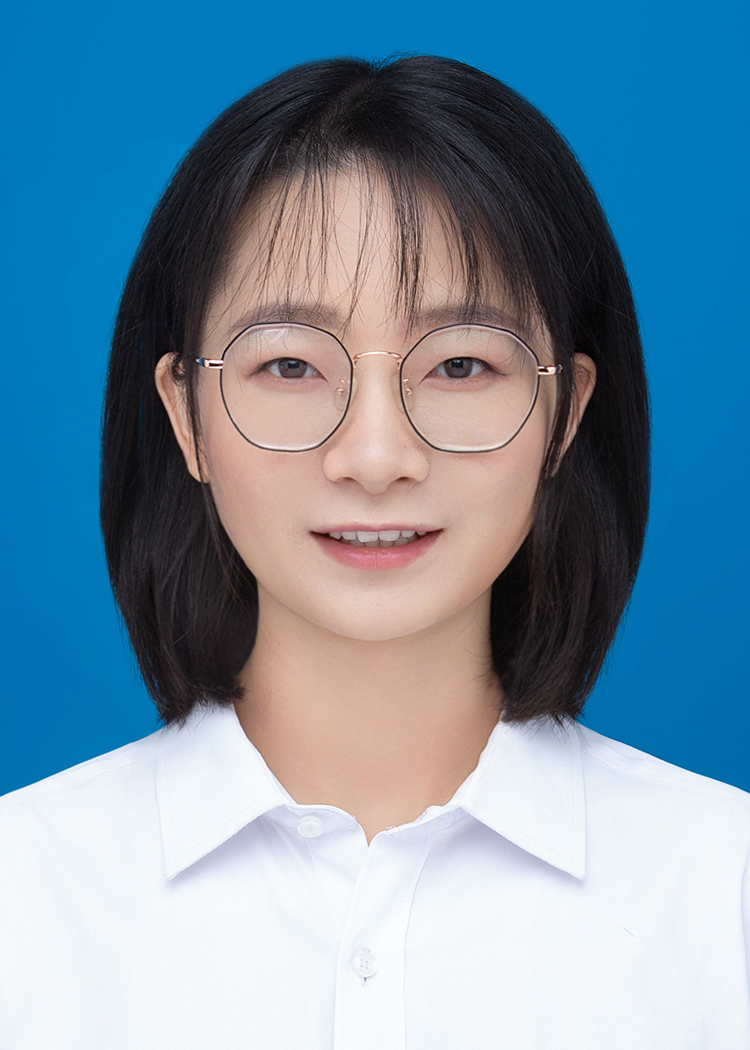}}]{Yuqian Fu} is currently a postdoc researcher at INSAIT, Bulgaria. Previously, she worked as a postdoc researcher at Computer Vision Lab (CVL), ETH Zürich, Switzerland. She received her Ph.D. degree from the School of Computer Science, Fudan University, China, in June 2023. Her research topics are vision and deep learning, especially transfer learning, domain adaptation, and multimodal learning. 
\end{IEEEbiography}

\begin{IEEEbiography}[{\includegraphics[width=1in,height=1.25in,clip,keepaspectratio]{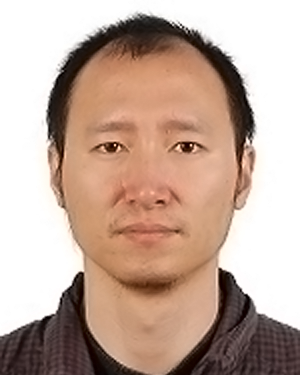}}]{Yun Wang} received the B.S. and M.S. degrees in electrical engineering from Wuhan University, Hubei Province, China, in 2011 and is currently pursuing a Ph.D. degree in the Institute of Science and Technology for Brain-Inspired Intelligence at Fudan University. From 2011 to 2017, he was a Research Engineer in State Grid Electric Power Research Institute. His current research interests include computational neuroscience and brain-inspired intelligence.
\end{IEEEbiography}

\begin{IEEEbiography}[{{\includegraphics[clip,width=1in,height=1.25in,keepaspectratio]{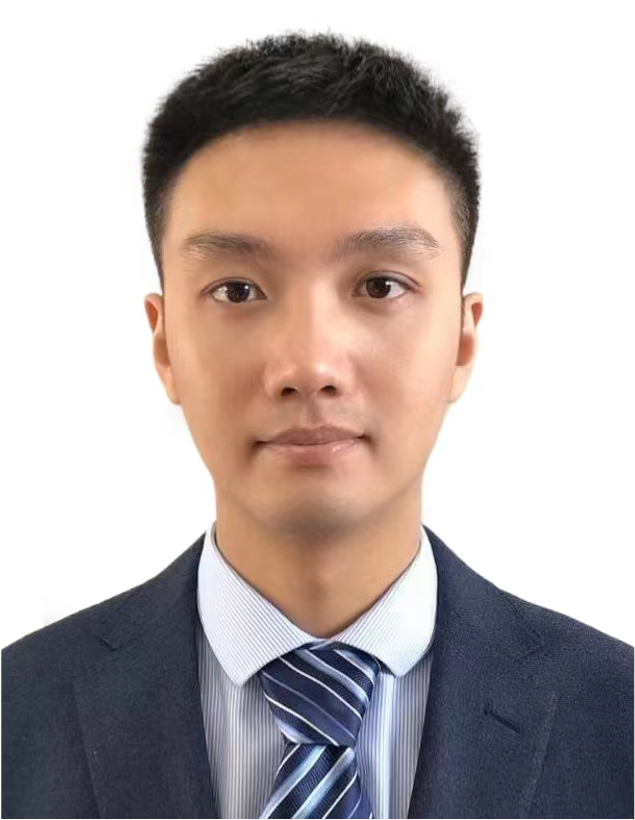}}}]{Xuelin Qian} (Member, IEEE) is an Associate Professor in the School of Automation, Northwestern Polytechnical University (NWPU). Before that, he held a post-doctoral position with Fudan University from 2022 to 2024. He received the Ph.D. degree from Fudan University in 2021, and the B.S. degree from Xidian University in 2015. He has published over 15 papers in top-tier conferences and journals, and served as a reviewer for CVPR, ICCV, TPAMI, IJCV \textit{etc}. His research interests are image retrieval, multi-modal generation, and medical image analysis.
\end{IEEEbiography}

\begin{IEEEbiography}[{\includegraphics[width=1in,height=1.25in,clip,keepaspectratio]{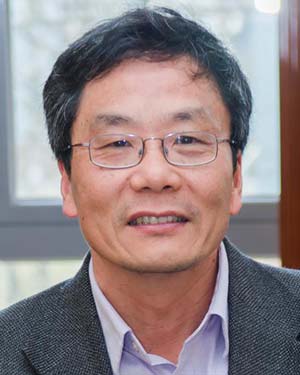}}]{Jianfeng Feng} (Senior Member, IEEE) received the
BS, MS, and PhD degrees from the Department of Probability and Statistics, Peking University, China. He is the chair professor with the Shanghai National Centre for Mathematic Sciences and the dean with the Brain-Inspired AI Institute, Fudan University. He leads the DTB project. He has been developing new mathematical, statistical, and computational theories and methods to meet the challenges raised in neuroscience and mental health research.
\end{IEEEbiography}






\end{document}